\documentclass[journal,twocolumn]{IEEEtran}

\usepackage{amsmath}
\usepackage{caption}
\usepackage{algorithm}
\usepackage{algpseudocode}
\usepackage{setspace}
\usepackage{color,hyperref}
\usepackage{subcaption}
\usepackage{graphicx}
\usepackage{amsmath,amssymb} 
\usepackage{epsfig}
\usepackage{amsmath}
\usepackage{amssymb}
\usepackage{colortbl}
\usepackage{soul}
\usepackage{multirow}

\usepackage{pifont}
\usepackage{xcolor}
\usepackage{array}
\newcommand{\xmark}{\ding{55}}
\newcommand{\rmark}{\ding{52}}

\newcolumntype{I}{!{\vrule width 1.2pt}}
\newlength\savedwidth
\newcommand\whline{\noalign{\global\savedwidth\arrayrulewidth
		\global\arrayrulewidth 1.25pt}%
	\hline
	\noalign{\global\arrayrulewidth\savedwidth}}

\usepackage{algorithm}
\usepackage{algorithmicx}

\ifCLASSINFOpdf
\else
\fi

\definecolor{darkblue}{rgb}{0.0,0.0,1.0}
\hypersetup{colorlinks,breaklinks,
	linkcolor=darkblue,urlcolor=darkblue,
	anchorcolor=darkblue,citecolor=darkblue}

\begin{document}
	
	\setul{}{1.5pt}
	
	\title{Feature-aware Adaptation and Density Alignment for Crowd Counting in Video Surveillance}
	
	\author{Junyu~Gao,\IEEEmembership{~Student Member,~IEEE}, ~Yuan~Yuan,\IEEEmembership{~Senior Member,~IEEE},  and Qi~Wang*,\IEEEmembership{~Senior Member,~IEEE}
		\thanks{* Q. Wang is the corresponding author.
			
			Junyu Gao, Yuan Yuan and Qi Wang are with the School of Computer Science and the Center for OPTical IMagery Analysis and Learning (OPTIMAL), Northwestern Polytechnical University, Xi'an 710072, China (e-mail: gjy3035@gmail.com; y.yuan1.ieee@gmail.com; crabwq@gmail.com). 
		
			\copyright 20XX IEEE. Personal use of this material is permitted. Permission from IEEE must be obtained for all other uses, in any current or future media, including reprinting/republishing this material for advertising or promotional	purposes, creating new collective works, for resale or redistribution to servers or lists, or reuse of any copyrighted component of this work in other works.	
}
	}
	\markboth{{IEEE} Transactions on Cybernetics}%
	{Shell \MakeLowercase{\textit{et al.}}: Bare Demo of IEEEtran.cls for Journals}
	\maketitle

\begin{abstract}
	With the development of deep neural networks, the performance of crowd counting and pixel-wise density estimation are continually being refreshed. Despite this, there are still two challenging problems in this field: 1) current supervised learning needs a large amount of training data, but collecting and annotating them is difficult; 2) existing methods can not generalize well to the unseen domain. A recently released synthetic crowd dataset alleviates these two problems. However, the domain gap between the real-world data and synthetic images decreases the models' performance. To reduce the gap, in this paper, we propose a domain-adaptation-style crowd counting method, which can effectively adapt the model from synthetic data to the specific real-world scenes. It consists of Multi-level Feature-aware Adaptation (MFA) and Structured Density map Alignment (SDA). To be specific, MFA boosts the model to extract domain-invariant features from multiple layers. SDA guarantees the network outputs fine density maps with a reasonable distribution on the real domain. Finally, we evaluate the proposed method on four mainstream surveillance crowd datasets, Shanghai Tech Part B, WorldExpo'10, Mall and UCSD. Extensive experiments evidence that our approach outperforms the state-of-the-art methods for the same cross-domain counting problem. 
\end{abstract}


\section{Introduction}

During the last half decade, Convolutional Neural Network(CNN)-based methods \cite{onoro2016towards,hu2016dense,babu2017switching,babu2018divide,shi2019revisiting,wan2019residual,liu2019crowd,8949751} attain a significant progress in the field of crowd counting, especially on some research datasets including some surveillance scenes, such as UCSD \cite{chan2008privacy}, Mall \cite{chen2012feature}, Shanghai Tech B \cite{zhang2016single}, WorldExpo'10 \cite{zhang2016data} and Zhengzhou Airport \cite{jiang2019learning}. Accurate counting is important to some high-level vision tasks, such as crowd analysis \cite{7434629,7165622,li2014crowded,yuan2018structured}, scene understanding \cite{8802265,zhao2019property,DBLP:conf/cvpr/Xiong0G020} and video analysis \cite{yuan2017tracking,6786497,zhao2019cam,DBLP:journals/ijon/XiongYW20}. However, current supervised CNN-based methods need a large amount of labeled data to train a counter network. Unfortunately, the aforementioned datasets contain only a small amount of samples so that the current CNN-based methods can not be adapted to the real world. To improve the performance in the wild, theoretically, labeling more data is a potential method. Nevertheless, collecting and annotating congested crowd scenes is a laborious and expensive process. Statistically, for a still image contains $\sim 500$ people, the labeling process takes about $30 \sim 40$ minutes, which is slower than other image- or object-level annotation tasks (such as image recognition \cite{deng2009imagenet,zhao2019weather} and object detection \cite{redmon2016you} or other machine learning tasks \cite{ijcai2020-416}).

\begin{figure}
	\centering
	\includegraphics[width=0.48\textwidth]{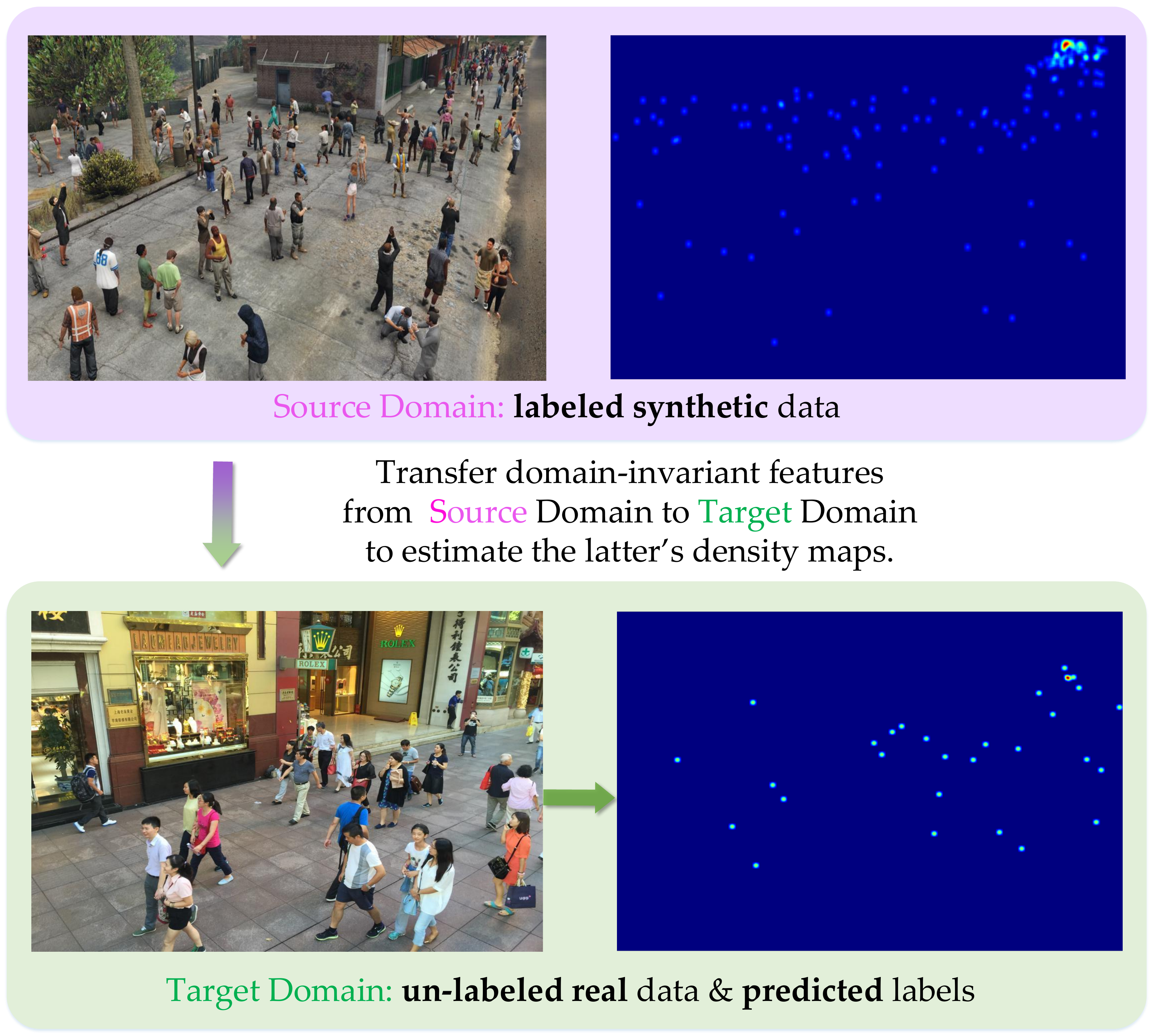}
	\caption{The goal of crowd counting via domain adaptation is: training a crowd counter using labeled synthetic data and applying it on the un-labeled real data. }\label{Fig-intro}
\end{figure}

Considering the expensive labeling costs, some researchers focus on dealing with the scarce labeled data from two aspects: data generation and methodology. For the former, Wang \emph{et al.} \cite{wang2019learning} construct a large-scale and synthetic GTAV Crowd Counting (GCC) dataset that automatically generated and labeled by a computer game mod. Unfortunately, the synthetic scenes are very different from the real world, of which difference is named as ``domain gap''. It results in a performance reduction when transferring a counter from the synthetic domain to the real-world domain. 

From the perspective of methodology, Liu \emph{et al.} \cite{liu2018leveraging} propose a learning-to-rank framework via leveraging unlabeled data. By this strategy, they exploit a large amount of unlabeled data to aid supervised learning. Sam \emph{et al.} \mbox{\cite{sam2019unsupervised}} present an almost unsupervised autoencoder for dense crowd counting, whose 99.9\% parameters are trained without any labeled data. However, these methods still rely on manually labeled data to a different extent.

For handling the performance reduction from GCC to real data, Wang \emph{et al.} \cite{wang2019learning,wang2020pixel} propose a domain adaptation method via SE Cycle GAN. Fig. \ref{Fig-intro} demonstrates this problem, i.e., how to exploit free synthetic labeled data to prompt the counting performance on real-world data. SE Cycle GAN translates the synthetic image to the photo-realistic image, which is a visual and explicit adaptation. Then they train a CNN model on translated images, which performs better on real data than the CNN models without domain adaptation. However, the translated images lose some detailed information (including texture, sharpness and local structure), especially in the congested crowd regions. 

Inspired by SE Cycle GAN \cite{wang2019learning}, this paper focuses on domain-adaptation-style crowd counting. Different from SE Cycle GAN's explicit translation, we propose a feature-aware adaptation to reduce the domain gap at feature level and output a reasonable structured density map. It consists of a common VGG-backbone Crowd Counter (CC), Multi-level Feature-aware Adaptation (MFA) and a Structured Density map Alignment (SDA). Since the design of CC is not the core of this paper, a state-of-the-art model (SFCN \cite{wang2019learning}) is only adopted. MFA aims to extract domain-invariant features from CC, which constructs two element-wise discriminators to classify the source (from synthetic or real data) of the feature maps extracted from different layers in CC. By adversarial learning, the feature maps in CC can confuse the discriminators, so the domain gap in the feature space is effectively alleviated. 

\begin{figure*}[t]
	\centering
	\includegraphics[width=1\textwidth]{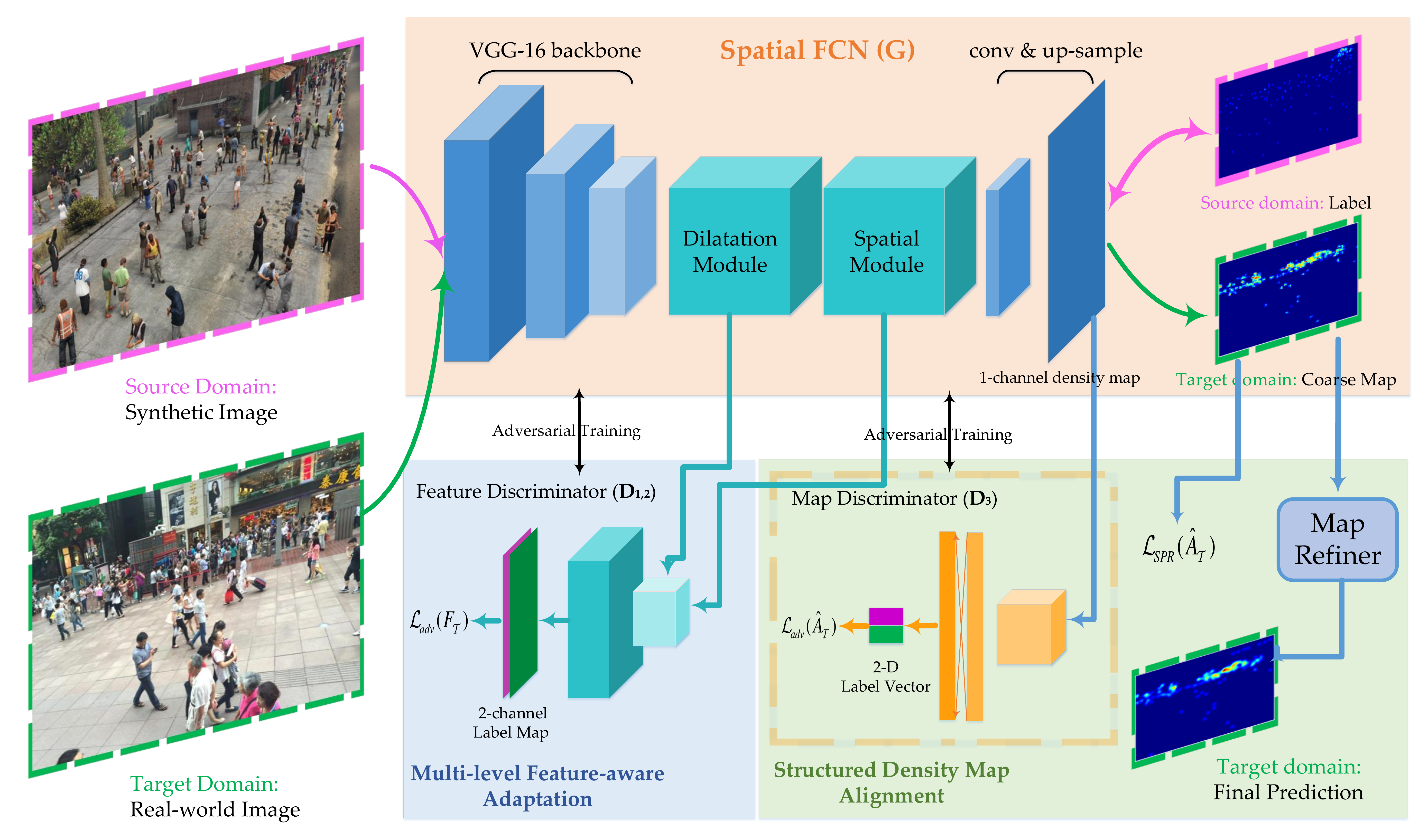}
	\caption{Spatial FCN (SFCN) is trained on the source domain and it directly estimates the density maps on the target domain. The bottom left box demonstrates Multi-level Feature-aware Adaptation (MFA), which classifies the pixel-wise label of feature maps. The bottom right describes Structured Density map Alignment (SDA), which consists of Map Discriminator (MD), Self-supervised Pyramid Residual (SPR) learning and Map Refiner (MR). By iteratively optimizing SFCN and the domain classifiers (MFA and MD), the final SFCN can extract the domain-invariant features.} \label{Fig-overview}
\end{figure*}

When introducing MFA, however, the quality of the density map is not good for the unseen real data. The concrete problems are: 1) some abnormal values exist in the map, 2) the map is very coarse and only reflects the density trend. To remedy these problems, we present a Structured Density map Alignment (SDA), which consists of three components. The first is similar to MFA: it is a Map Discriminator (MD) to classify the source of the density map output by CC. The second is a Self-supervised Pyramid Residual (SPR) learning on the target domain, which can maintain the consistency of density maps at different scales. The last component is named as ``Map Refiner'', which receives the coarse density map of CC then outputs the fine and reasonable map. Fig. \ref{Fig-overview} shows the flowchart of our proposed crowd counting via domain adaptation.

As a summary, the contributions of this paper are:
\begin{enumerate}
	\item[1)] To our best knowledge, this paper is the first to propose a feature-aware adaptation for crowd counting. 
	\item[2)] This paper designs a Structured Density map Alignment (SDA) to refine the quality of density maps for unseen scenes.
	\item[3)] The proposed method yields a new record of MAE and MSE on the domain-adaptation-style crowd counting from synthetic to real data.	
\end{enumerate}

\section{Related Works}

Here, we briefly review the mainstream works about the two most related tasks: crowd counting and domain adaptation from synthetic to real data. 

\subsection{Crowd Counting}

\subsubsection{Traditional Supervised Learning}

Many algorithms are proposed to handle crowd counting in the last decade. \cite{kong2006viewpoint,rabaud2006counting,chen2012feature} adopt some hand-crafted features to train the counting regressors, such as HOG, SIFT and so on. With the development of CNN, a large number of CNN-based methods \cite{zhang2016single,sindagi2017generating,sam2018top,li2018csrnet,cao2018scale,ranjan2018iterative,gao2019scar}  attain the significant performance. Zhang \emph{et al.} \cite{zhang2016single} propose a multi-column CNN to cover the respective fields with different sizes for the images. To encode more contextual information, CP-CNN is presented by \cite{sindagi2017generating}, which uses a global/local context networks to assist the counting. Considering that CP-CNN's complex training scheme, Li \emph{et al.} \cite{li2018csrnet} propose a single CNN model (CSRNet) to encode more large-range features. 
Cao \emph{et al.} \cite{cao2018scale} propose a simple and efficient Scale Aggregation Network (SANet) to output structured density maps. To further explore the deep features, some methods \cite{idrees2018composition,liu2019context,sindagi2019multi,jiang2019crowd} focus on feature fusion to improve counting. In addition to directly regressing the density map, some researchers introduce other auxiliary tasks to prompt the counting performance \cite{sindagi2017cnn,gao2019pcc,zhao2019leveraging,he2019dynamic,Lian_2019_CVPR,wan2020fine}.

\subsubsection{Counting for Scarce Labeled Data}

Due to the high cost of manually labeling data, some works \cite{Elassal2016Unsupervised,liu2018leveraging,sam2019unsupervised,wang2019learning} attempt to tackle this problems. Elassal and Elder \cite{Elassal2016Unsupervised} are the first to propose an unsupervised crowd counting via automatically learning how many groups of people. However, it cannot perform well in the congested crowd scenes. Liu \emph{et al.} \cite{liu2018leveraging} exploit lots of unlabeled crowd scenes to reduce the estimation errors of traditional supervised learning. Sam \emph{et al.} \cite{sam2019unsupervised} propose an almost unsupervised autoencoder for crowd counting, of which only 0.1\% parameters need labeled data during learning process. To completely get rid of manually labeled data, Wang \emph{et al.} \cite{wang2019learning} construct a synthetic GCC dataset and propose a domain-adaptation-style method, which does not need any labeled real data. 
At the same time, \cite{wang2019learning} presents a novel training scheme: the model is firstly pre-trained on GCC and then fine-tuned on real-world datasets. By this strategy, the model performs better than traditional training without preliminary training on GCC.

\subsection{Domain Adaptation}

For exploiting synthetic data to prompt the classification performance on real data, some methods \cite{ganin2016domain,tzeng2017adversarial,wen2019exploiting,wen2019bayesian,gao2020nwpu} attempt to reduce the domain gap. With the release of some synthetic segmentation datasets \cite{richter2016playing,ros2016synthia}, many researchers pay attention to the task of pixel-wise domain adaptation \cite{hoffman2016fcns,hoffman2017cycada,sankaranarayanan2018learning,gao2019weakly,lee2018diverse,chen2019crdoco,8784811}. Hoffman \emph{et al.} \cite{hoffman2016fcns} firstly propose an unsupervised domain adaptation for semantic segmentation, including the global and category adaptation. Sankaranarayanan \emph{et al.} \cite{sankaranarayanan2018learning} propose a joint adversarial learning approach, which transfers the target distribution to the learned embedding. Hoffman \emph{et al.} \cite{hoffman2017cycada} present a Cycle-Consistent Adversarial Domain Adaptation (CyCADA) for unsupervised semantic segmentation based on \cite{zhu2017unpaired}. 
For producing structured segmentation masks, \cite{tsai2018learning} designs a multi-level adversarial network to reduce the domain gap between synthetic and real data. To remedy the intrinsic differences in different domains, Chen \emph{et al.} \cite{chen2018road} propose a spatial-aware adaptation scheme to align the feature distribution of two domains. Benefiting from the \cite{zhu2017unpaired}, Chen \emph{et al.} \cite{chen2019crdoco} present CrDoCo for domain-adaptive dense prediction tasks, which contains two steps: 1) a CycleGAN-style image translation and 2) two task networks for the specific domain. Wang \mbox{\emph{et al.} \cite{8784811}} propose a two-stage density adaptation method, consisting of training on the source domain and adversarial learning on the target domain.

\section{Methodology}

\subsection{Algorithmic Overview}

The proposed domain-adaptation-style crowd counting consists of three modules: 1) Spatial FCN: a counter network ($\boldsymbol{G}$); 2) Multi-level Feature-aware Adaptation (MFA): two domain discriminators ($\boldsymbol{D_1}$, $\boldsymbol{D_2}$); 3) Structured Density map Alignment (SDA): a map domain discriminator ($\boldsymbol{D_3}$) and a map refiner ($\boldsymbol{R}$). The data include: source domain images ${I_\mathcal{S}}$, source domain labels $A_\mathcal{S}$ and target domain images ${I_\mathcal{T}}$, where ${\mathcal{S}}$ and ${\mathcal{T}}$ indicates the source and target domain, respectively.

Based on the above symbols, the goal of this paper is described as three following steps: 

\begin{enumerate}
	\item[1)] Given ${I_\mathcal{S}}$,  $A_\mathcal{S}$ and ${I_\mathcal{T}}$, by the supervised learning on ${\mathcal{S}}$ for $\boldsymbol{G}$ and the adversarial learning for $\boldsymbol{G}$ and $\boldsymbol{D_i}$ ($i = 1,2,3$), $\boldsymbol{G}$ can predict coarse maps $\hat A_\mathcal{T}$ of ${I_\mathcal{T}}$.
	\item[2)] Using ${I_\mathcal{S}}$,  $A_\mathcal{S}$, a synthetic counter network is trained to predict the density map $\hat A_\mathcal{S}$. Then, a map refiner $\boldsymbol{R}$ is trained using $\hat A_\mathcal{S}$ and $A_\mathcal{S}$, of which $\hat A_\mathcal{S}$ are inputs and $A_\mathcal{S}$ are labels during training process. 
	\item[3)] Given the coarse map $\hat A_\mathcal{T}$ in Step(1) as inputs, the trained $\boldsymbol{R}$ in Step(2) outputs final maps $\hat A_\mathcal{T}^{final}$.
\end{enumerate}

Here, we formulate the loss functions during the training in Step(1) and (2). Specifically, the former is defined as:

\begin{equation}
\begin{array}{l}
\begin{aligned}
\mathcal{L}\left( I_{\mathcal{S}},A_{\mathcal{S}},I_{\mathcal{T}} \right) =& {\mathcal{L}_{cnt}}({I_{\mathcal{S}}},{A_{\mathcal{S}}}) + \lambda {\mathcal{L}_{adv}}(F_{\mathcal{T}}) \\&+ \beta {\mathcal{L}_{adv}}(\hat A_{\mathcal{T}}) + \gamma \mathcal{L}_{SPR}(\hat A_{\mathcal{T}}),
\end{aligned}
\end{array}\label{all_loss}
\end{equation}
where $\mathcal{L}_{cnt}$ is the standard MSE loss, $\mathcal{L}_{adv}$ is the adversarial loss, and $\mathcal{L}_{SPR}$ is self-supervised pyramid residual loss. $\lambda$, $\beta$ and $\gamma$ are the weights to balance the losses. The concrete descriptions about $\mathcal{L}_{adv}$ and $\mathcal{L}_{SPR}$ are explained in the next section. The loss function of Step(2) is the standard MSE.

\subsection{Multi-level Feature-aware Adaptation}
\label{MFA}

Since there are domain gaps between two different domains, the counter network trained by traditional supervised learning on a specific domain can not perform well on other domains. Thus, it is important that how to reduce the impact of domain gaps during the training. In other words, the purpose of adaptation is to improve the counter network to extract domain-invariant features. To this end, we present the multi-level feature-aware adaptation to reduce the domain gap of feature maps in the networks.

Since the crowd counting (density estimation) is a pixel-wise regression problem, a domain discriminator is designed to classify each unit of feature maps. To be specific, it is a fully convolutional network, including four convolution layers with leaky ReLU. It outputs a 2-channel score map with the same size as the input feature map. The values of each channel represent the confidence belonging to the source or target domain.

For the feature maps (${F_{\mathcal{S}}^i}$, ${F_{\mathcal{T}}^i}$, $i=1,2$) of Dilatation and Spatial Module in $\boldsymbol{G}$, we train two discriminators for them. Through $\boldsymbol{D_i}$, the pixel-wise domain labels $O_\mathcal{S}^{{D}}$ and $O_\mathcal{T}^{{D}}$ can be obtained. For optimizing $\boldsymbol{D_i}$, we adopt 2-D pixel-wise binary cross-entropy loss as the objective function, which is formulated as:
\begin{equation}
\begin{array}{l}
\begin{aligned}
\mathcal{L}_D^i({F_{\mathcal{S}}^i},{F_{\mathcal{T}}^i}) = & - \sum\limits_{F_\mathcal{S}^i \in \mathcal{S}} {\sum\limits_{h \in H} {\sum\limits_{w \in W} {\log (p(O_\mathcal{S}^{D})) } } } \\
&- \sum\limits_{F_\mathcal{T}^i \in \mathcal{T}} {\sum\limits_{h \in H} {\sum\limits_{w \in W} {\log (1 - p(O_\mathcal{T}^{D}))} } } ,
\end{aligned}\label{L_D}
\end{array}
\end{equation} 
where $O_\mathcal{S}^{D}$ and $O_\mathcal{T}^{D}$ are 2D-channel predicted maps, with size of $H \times W$ corresponding to the source and target inputs, and $p( \cdot )$ is a standard soft-max operation at the pixel level. 

In order to extract domain-invariant features to confuse $\boldsymbol{D_i}$, the inverse adversarial loss should be added into the training process of $\boldsymbol{G}$, which is defined as:
\begin{equation}
\begin{array}{l}
{\mathcal{L}_{adv}^i(F_{\mathcal{T}}^i)} = - \sum\limits_{i = 1}^2 {\sum\limits_{F_\mathcal{T} \in \mathcal{T}} {\sum\limits_{h \in H} {\sum\limits_{w \in W} {\log (p(O_\mathcal{T}^{D}))} } }}, i=1,2.
\label{adv}
\end{array}
\end{equation}
This loss guides $\boldsymbol{G}$ to fool the two discriminators $\boldsymbol{D_1}$, $\boldsymbol{D_2}$. Thus, the domain gaps at different feature levels are effectively alleviated.

\subsection{Structured Density Map Alignment}
\label{SDA}
Although the domain gap between different domains is alleviated to some extent, however, there is an intractable issue to be tackled, i.e., the quality of the density map is not good for unseen data. To be specific, the map contains some abnormal regression results, and the map is coarse so that it only reflects the density trend. To this end, we propose a structured density map alignment approach to refine the coarse maps.

\subsubsection{Map Discriminator}

Other domain adaptation tasks \cite{hoffman2016fcns,hoffman2017cycada,tsai2018learning} (such as image classification and segmentation) usually only focus on constructing domain discriminators at the feature level. The main reason is that: in the aforementioned tasks, each image/pixel is assigned with a specific category by soft-max or other classification layers. In other words, there is no invalid value in the prediction map. However, as for crowd counting, the regression model may produce some abnormal values on unseen domains. 


To handle this problem, a map discriminator (MD) $\boldsymbol{D_3}$ is proposed to classify whether the predicted density maps from source or target domain. MD consists of three convolution layers with leaky ReLU and a fully-connected layer. It receives 1-channel prediction maps $\hat A_{\mathcal{S}}$ and $\hat A_{\mathcal{T}}$, then produces 2-D score vectors $V_{\mathcal{S}}$ and $V_{\mathcal{T}}$, which denote the confidence belonging to source or target domain. For training $\boldsymbol{D_3}$, the loss function as below is optimized:   
\begin{equation}
\begin{array}{l}
\begin{aligned}
\mathcal{L}_D^3({\hat A_{\mathcal{S}}},{\hat A_{\mathcal{T}}}) = &  - \sum\limits_{\hat A_\mathcal{S} \in \mathcal{S}} {\log (p(V_\mathcal{S})) } \\
& - \sum\limits_{\hat A_\mathcal{T} \in \mathcal{T}} {  {\log (1 - p(V_\mathcal{T}))} } ,
\end{aligned}\label{L_D3}
\end{array}
\end{equation}
where $p( \cdot )$ is the soft-max operation. In fact, Eq. \ref{L_D3} is treated as a binary cross-entropy loss. Like MFA, the inverse loss is introduced into $\boldsymbol{G}$ to fool $\boldsymbol{D_3}$, which is defined as:  

\begin{equation}
\begin{array}{l}
{\mathcal{L}_{adv}(\hat A_{\mathcal{T}})} = - \sum\limits_{\hat A_\mathcal{T} \in \mathcal{T}} { \log (p(V_\mathcal{T})) }.
\label{adv_D3}
\end{array}
\end{equation}

\subsubsection{Self-supervised Pyramid Residual Learning}

Theoretically, resizing images with a similar size does not result in dramatic changes in image content. Thus, the counter is supposed to estimate the different but close number of people.  However, the counter is not very robust under this circumstance. Therefore, we propose a Self-supervised Pyramid Residual (SPR) learning method to remedy the above problem. 

To be specific, given a standard target image $i_\mathcal{T}^{1.0x} \in I_\mathcal{T}$, the counter $\boldsymbol{G}$ produce a density map $\hat a_\mathcal{T}^{1.0x}$. At the same time, $i_\mathcal{T}^{1.0x}$ is resize to a smaller and a larger image, namely $i_\mathcal{T}^{mx}$ ($0.8<m<1.0$) and $i_\mathcal{T}^{nx}$ ($1.0<n<1.2$), respectively. As a results, the three predicted density maps are obtained, namely $\hat a_\mathcal{T}^{1.0x}$, $\hat a_\mathcal{T}^{mx}$, and $\hat a_\mathcal{T}^{nx}$. To maintain the consistency of density map at the pixel level, we need to semantically reshape $\hat a_\mathcal{T}^{yx}$ ($y=m,n$) to the resolution with $1.0x$. As for density maps, the semantic reshaping of $px \to qx$ ($p,q>0$) is defined approximately as follows:
\begin{equation}
\begin{array}{l}
{a^{px}} \Leftrightarrow r({a^{qx \to px}}) \times \dfrac{q}{p} \times \dfrac{q}{p},
\label{reshape}
\end{array}
\end{equation}
where $r( \cdot )$ is the image resizing operation. Finally, the loss of SPR is formulated as:
\begin{equation}
\begin{array}{l}
\mathcal{L}_{SPR}(\hat A_{\mathcal{T}}) = \sum\limits_{y = m,n} {MSE[{\hat a_\mathcal{T}^{1.0x}},p({\hat a_\mathcal{T}^{yx \to 1.0x}}) \times {y^2}]},
\label{loss_SPR}
\end{array}
\end{equation}
where $MSE[ \cdot ]$ is the standard Mean Squared Error.

Note that we set $m$ and $n$ as a value close to $1$, which guarantees the image content change not much. In practice, they are randomly set in $0.8, 1.0$ and $1.0, 1.2$, respectively.

\subsubsection{Map Refiner}
\label{mr}

As for the problem of coarse density maps, a Map Refiner (MR) is presented to improve the map quality. According to the MFA experiments, the producing coarse maps is very common in the cross-domain crowd counting. In order to handle this problem, we attempt to simulate the phenomenon using synthetic data and then only exploit the coarse map to reconstruct the fine map. 

\begin{figure}
	\centering
	\includegraphics[width=0.5\textwidth]{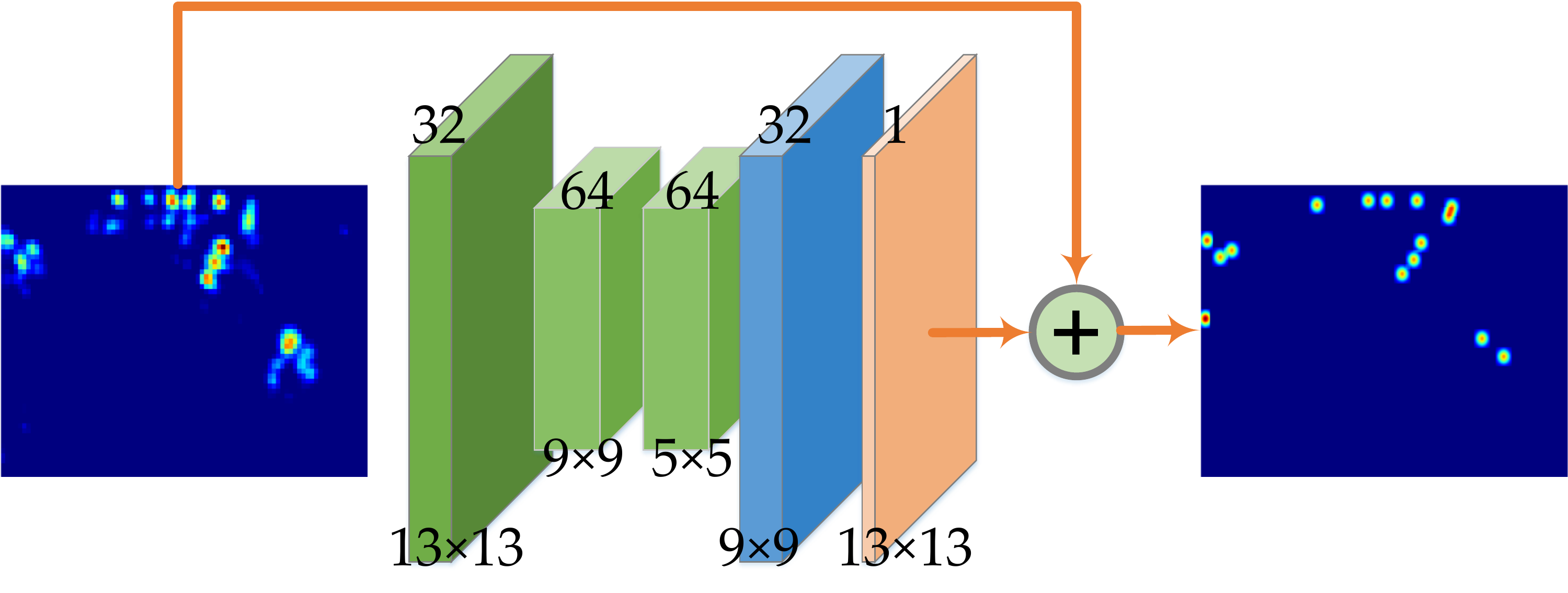}
	\caption{The map flow of the proposed MR. }\label{Fig-MRnet}
\end{figure}

To be specific, in GCC dataset, the authors of \cite{wang2019learning} provide three evaluation schemes to assess the generalization ability of the model. Here, we select the cross-location splitting to mimic the cross-domain problem. In this scheme, training and testing are conducted the crowd scenes in different locations of the GTA V world, of which data are very different. Thus, the cross-location evaluation also suffers from the problem of coarse maps. After SFCN is successfully trained on GCC, it is applied to the test set and produces the prediction map. At the same time, the original test data are randomly split into training/validation/testing for modeling MR according to the ratio of 70/10/20\%. The MR consists of three convolutional layers, a de-convolutional layer, and a regression layer. Except for the regression layer, we adopt PReLU to restrict the output for each layer. The concrete architecture is illustrated in Fig. \ref{Fig-MRnet}. The final output is the sum of the regression result and the input, which aims to maintain the key original density map.

In the design of MR, the large kernels (such as $13 \times 13$ and $9 \times 9$) are used to cover a larger respective field. As a result, MR can effectively reduce the low-response estimation noises and aggregate the high-response region. From the whole architecture, our goals are: 1) Design a light network to reconstruct a finer density map. So, we construct a five-layer network. 2) To maintain a high-resolution spatial feature map, we only conduct three convolutional layers and down-sample the feature map to 0.5x. Then the de-convolutional and regression layer is added to output the residual map with the size of the original input density map. 

The refinement of MR is illustrated in Algorithm \ref{alg1}.

\begin{algorithm}[htb] 
	\caption{ Algorithm for refinement pipeline of MR.} 
	\begin{algorithmic}[1]
		\Require Training data ${Tr_\mathcal{S}}$ and testing data ${Te_\mathcal{S}}$ in $\mathcal{S}$, coarse predicted maps $\hat A_\mathcal{T}$ in $\mathcal{T}$;
		\Ensure The map refiner $\boldsymbol{R}$ and the refined maps $\hat A_\mathcal{T}^{final}$.
		\State Train a counter network $\boldsymbol{G}_\mathcal{S}$ on ${Tr_\mathcal{S}}$;
		\State Perform $\boldsymbol{G}_\mathcal{S}$ on ${Te_\mathcal{S}}$;
		\State Split ${Te_\mathcal{S}}$ into training/validation/testing set: ${Te_\mathcal{S}^{tr}}$, ${Te_\mathcal{S}^{val}}$, ${Te_\mathcal{S}^{te}}$;
		\State Train a map refiner $\boldsymbol{R}$ on ${Te_\mathcal{S}^{tr}}$, ${Te_\mathcal{S}^{val}}$ using standard MSE Loss;
		\State Apply $\boldsymbol{R}$ on $\hat A_\mathcal{T}$ to obtain $\hat A_\mathcal{T}^{final}$;\\
		\Return $\boldsymbol{R}$ and $\hat A_\mathcal{T}^{final}$. 
	\end{algorithmic}\label{alg1}
\end{algorithm}

\section{Implementation}

\subsection{Iterative optimization}

During the training phase, we adopt the iterative optimization strategy to train $\boldsymbol{G}$, and $\boldsymbol{D_i}$ ($i = 1,2,3$). Another network $\boldsymbol{R}$ is trained independently. The iterative optimization is explained as follows:

\textbf{(1) G-update}: Fix the parameter of $\boldsymbol{D_i}$ ($i = 1,2,3$), and update the parameter of $\boldsymbol{G}$ by optimizing Eq. \ref{all_loss};

\textbf{(2) D-update}: Fix the parameter of $\boldsymbol{G}$, and update the parameter of $\boldsymbol{D_i}$ ($i = 1,2,3$) by respectively optimizing Eq. \ref{L_D} and \ref{L_D3}.

By repeating the Step (1) and (2), the anti-domain-gap $\boldsymbol{G}$ is obtained.

The entire pipeline of training process is describe as below:

\begin{algorithm}[htb] 
	\caption{ Algorithm for the training process of the entire architecture.} 
	\begin{algorithmic}[1]
		\Require The initialized Counter $\boldsymbol{G}$ and the initialized Discriminator $\boldsymbol{D_i}$ ($i = 1,2,3$);
		\Ensure The trained Counter $\boldsymbol{G}$.
		\State \textbf{repeat}
		\State \,\,\, Compute the loss $\mathcal{L}_D^i$ of $\boldsymbol{D_i}$ ($i = 1,2,3$) and the corresponding the gradient; 
		\State \,\,\, Back-propagate gradients and update the parameter of $\boldsymbol{D_i}$ ($i = 1,2,3$); 
		\State \,\,\, Compute the loss $\mathcal{L}$ of $\boldsymbol{G}$ and the corresponding the gradient; 
		\State \,\,\, Back-propagate gradients and update the parameter of $\boldsymbol{G}$;  
		\State \textbf{until} the task loss $\mathcal{L}_{cnt}$ is increased on the validation data.\\
		\Return Select the best Counter $\boldsymbol{G}$ on the validation data. 
	\end{algorithmic}\label{alg2}
\end{algorithm}

\subsection{Parameter Setting}
This section reports the key parameter setting in the experiments. During the adversarial training, the $\lambda$, $\beta$ and $\gamma$ in Eq. \ref{all_loss} are set as ${10^{ - 3}}$, ${10^{ - 3}}$ and ${10^{ - 1}}$, respectively. The learning rates of $\boldsymbol{G}$ and $\boldsymbol{D_i}$ ($i = 1,2,3$) are initialized at ${10^{ - 5}}$ and the $\boldsymbol{R}$'s learning rate is set as ${10^{ - 4}}$.  Adam \cite{kingma2014adam} algorithm is performed to optimize each network and obtain the best results. The training and evaluation are performed on NVIDIA GTX 1080Ti GPU using PyTorch Crowd Counting Framework \cite{gao2019c,steiner2019pytorch}.

\subsection{Scene Regularization}

Since GCC contains the crowd scenes under some special weathers or environments, training on the entire dataset may cause negative adaptation. To remedy the side effects, the Scene Regularization \cite{wang2019learning} is exploited to select the proper scenes. To be specific, we fully follow \cite{wang2019learning}'s setting on Shanghai Tech Part B \cite{zhang2016single} and WorldExpo'10 \cite{zhang2016data}. The selection settings on Mall \cite{chen2012feature} and UCSD  \cite{chan2008privacy} are described in Table \ref{Table-filter}.

\begin{table*}[htbp]
	
	\centering
	\caption{Filter condition on the four real datasets.}
	\begin{tabular}{c|c|c|c|c|c}
		\whline
		Target Dataset  & level & time & weather & count range & ratio range\\
		\whline
		Shanghai Tech Part B   &1,2,3,4,5 & 6:00$\sim$19:59 & 0,1,5,6 &10$\sim$600 & 0.3$\sim$1 \\
		\hline
		WorldExpo'10  &2,3,4,5,6 & 6:00$\sim$18:59 & 0,1,5,6 &0$\sim$1000 & 0$\sim$1\\
		\hline
		Mall   &1,2,3,4 & 8:00$\sim$18:59 & 0,1,5,6 &0$\sim$200 & 0$\sim$1 \\
		\hline
		UCSD  &1,2,3,4 & 8:00$\sim$18:59 & 0,1,5,6 &0$\sim$200 & 0$\sim$1\\
		\whline		
	\end{tabular}
	\label{Table-filter}
\end{table*}

\begin{table*}[htbp]
	\centering
	\caption{The performance of No Adaptation (NoAdapt), Cycle GAN, SE Cycle GAN, FSC and our approaches on the four real-world datasets.}
	\setlength{\tabcolsep}{2.0mm}{
		\begin{tabular}{c|cIc|cIc|c|c|c|c|cIc|cIc|c}
			\whline
			\multirow{2}{*}{Method}	&\multirow{2}{*}{DA} &\multicolumn{2}{cI}{SHT B} & \multicolumn{6}{cI}{WorldExpo'10 (MAE)} &\multicolumn{2}{cI}{Mall} &\multicolumn{2}{c}{UCSD}\\
			\cline{3-14} 
			& & MAE &MSE &S1 &S2 &S3 &S4 &S5 &Avg. & MAE &MSE & MAE &MSE \\
			\whline
			NoAdpt \cite{wang2019learning}  &\xmark &22.8 &30.6 &4.4 &87.2 &59.1  &51.8 &11.7 &42.8&-&-&-&- \\
			\hline
			Cycle GAN \cite{zhu2017unpaired} &\rmark &25.4 &39.7 &4.4 &69.6 &49.9 &29.2  &9.0 &32.4&-&-&-&- \\
			\hline	
			SE Cycle GAN \cite{wang2019learning} &\rmark&19.9 &28.3&4.3 &59.1 &43.7  &17.0 &7.6 &26.3 &-&-&-&-\\
			\hline
			SE Cycle GAN (JT) \cite{wang2020pixel} &\rmark&16.4 &25.8&\textbf{4.2} &\textbf{49.6} &41.3& 19.8 &7.2 &24.4 &-&-&-&-\\
			\hline
			FSC \cite{han2020focus} &\rmark&16.9 &\textbf{24.7}&\textbf{4.2} &54.7 &40.5  &\textbf{10.5} &36.4 &29.3 &3.71&4.66&3.85&4.90\\
			\whline
			NoAdpt (ours) &\xmark &22.3 &29.9 &5.4 &88.2 &62.1  &16.2 &14.3 &37.2 &4.07 &5.12&16.46&16.80\\
			\hline
			SFCN+MFA &\rmark &17.3 &26.1 &5.0 &71.4 &29.6&17.0 &7.0 &26.0 &2.87&3.57&2.39&2.91 \\
			\hline		
			SFCN+MFA+MD+SPR &\rmark&16.2 &24.9 &6.1 &60.4 &23.5  &16.1 &\textbf{6.8} &22.6 &2.56 &3.48&2.09 &2.51 \\
			\hline	
			SFCN+MFA+SDA &\rmark&\textbf{16.0} &\textbf{24.7}&5.7 &59.9 &\textbf{19.7}  &14.5 &8.1 &\textbf{21.6} &\textbf{2.47} &\textbf{3.25}&\textbf{2.00} &\textbf{2.43}  \\
			\whline			
		\end{tabular}
	}\label{Table-DA}
\end{table*}

The Arabic numerals in the above table are explained as follows:

\textbf{Level Categories} 0: 0$\sim$10, 1: 0$\sim$25, 2: 0$\sim$50, 3: 0$\sim$100, 4: 0$\sim$300, 5: 0$\sim$600, 6: 0$\sim$1k, 7: 0$\sim$2k and 8: 0$\sim$4k. 

\textbf{Weather Categories} 0: clear, 1: clouds, 2: rain, 3: foggy, 4: thunder, 5: overcast and 6: extra sunny. 

\textbf{Ratio range} is a restriction in terms of congestion. 
\section{Experiments}

\subsection{Metrics}

To evaluate the estimation performance of counting, the two mainstream criteria are used: Mean Absolute Error (MAE) and Mean Squared Error (MSE), which are formulated as:
\begin{equation}
\begin{array}{l}
MAE = \frac{1}{N}\sum\limits_{i = 1}^N {\left| {{y_i} - {{\hat y}_i}} \right|}, 
\label{metric1}
\end{array}
\end{equation}

\begin{equation}
\begin{array}{l}
MSE = \sqrt {\frac{1}{N}\sum\limits_{i = 1}^N {{{\left| {{y_i} - {{\hat y}_i}} \right|}^2}} },
\label{metric2}
\end{array}
\end{equation}
where $N$ is the number of testing samples, ${{y_i}}$ is the groundtruth label for counting and ${{{\hat y}_i}}$ is the estimated counting value for the $i$-th test sample. Besides, we adopt the Peak Signal-to-Noise Ratio (PSNR) and the Structural Similarity in Image (SSIM) \cite{wang2004image} to assess the quality of the predicted density maps.

\subsection{Datasets}
\label{dataset}
For evaluating the proposed method, we conduct the adaptation experiments from GCC Dataset \cite{wang2019learning} to another four real-world datasets containing the consistent crowd scene, namely Shanghai Tech Part B \cite{zhang2016single}, WorldExpo'10 \cite{zhang2016data}, Mall \cite{chen2012feature} and UCSD \cite{chan2008privacy}.

\textbf{GTA V Crowd Counting Dataset (GCC)}, a large-scale synthetic dataset, consists of still $15,212$ crowd images with resolution of $1080 \times 1920$, which are captured from 400 surveillance cameras in 100 locations of a fictional city. 

\textbf{Shanghai Tech Part B} \cite{zhang2016single} is collected from the surveillance camera on the Nanjing Road Pedestrian Street in Shanghai, China. It contains $400$ training and $316$ testing images with the same resolution of $768 \times 1024$. The entire dataset contains $88,488$ pedestrians.

\textbf{WorldExpo'10} is a cross-scene large-scale crowd counting dataset, which is present by Zhang \emph{et al.} \cite{zhang2016data}. All images are captured from 108 surveillance cameras in Shanghai 2010 WorldExpo, of which images from the 103 cameras are training data and the others are testing data. To be specific, WorldExpo'10 contains $3,980$ images with size of $576 \times 720$ and $199,923$ labeled pedestrians.

\textbf{Mall} \cite{chen2012feature} is an indoor crowd counting dataset, which is collected using a surveillance camera installed in a shopping mall. The dataset records the 2,000 sequential frames with resolution of $480 \times 640$. The first 800 frames are training samples and the others are test samples. 

\textbf{UCSD} \cite{chan2008privacy} is a single-scene dataset collected from a video camera at a pedestrian walkway. The dataset contains $2,000$ frames with a low resolution of $158 \times 238$, of which 601 to 1,400 images are training data and the others are testing samples.

\subsection{Adaptation to Real-world Datasets}

This section compares the proposed methods with other mainstream crowd counting via domain adaptation. All methods adopt Spatial FCN (SFCN) as a counter. Cycle GAN \cite{zhu2017unpaired} and SE Cycle GAN \cite{wang2019learning} translate synthetic scenes to photo-realistic scenes, then they train a counter network on translated images. Finally, the counter is applied to real-world datasets. Table \mbox{\ref{Table-DA}} lists the performance of No Adaptation (NoAdapt), Cycle GAN \mbox{\cite{zhu2017unpaired}}, SE Cycle GAN \mbox{\cite{wang2019learning}}, \mbox{FSC\cite{han2020focus}} and our approaches on the four real-world datasets. In order to show the real improvement, we also re-implement the SFCN without adaptation (NoAdpt) using our code, of which results are close to NoAdpt by \cite{wang2019learning}. Note that the results on WorldExpo'10 only report MAE and the final performance is the average of MAEs on five scenes. From the entire table, our proposed full model (SFCN+MFA+SDA, namely SFCN+MFA+MD+SPR+MR) outperforms other state-of-the-art methods on all datasets. Notably, compared with FSC that uses GAN for feature maps in one layer, the proposed method adopts GAN for multiple layer, which makes model extract more domain-invariant features.

\begin{figure*}[t]
	\centering
	\includegraphics[width=1\textwidth]{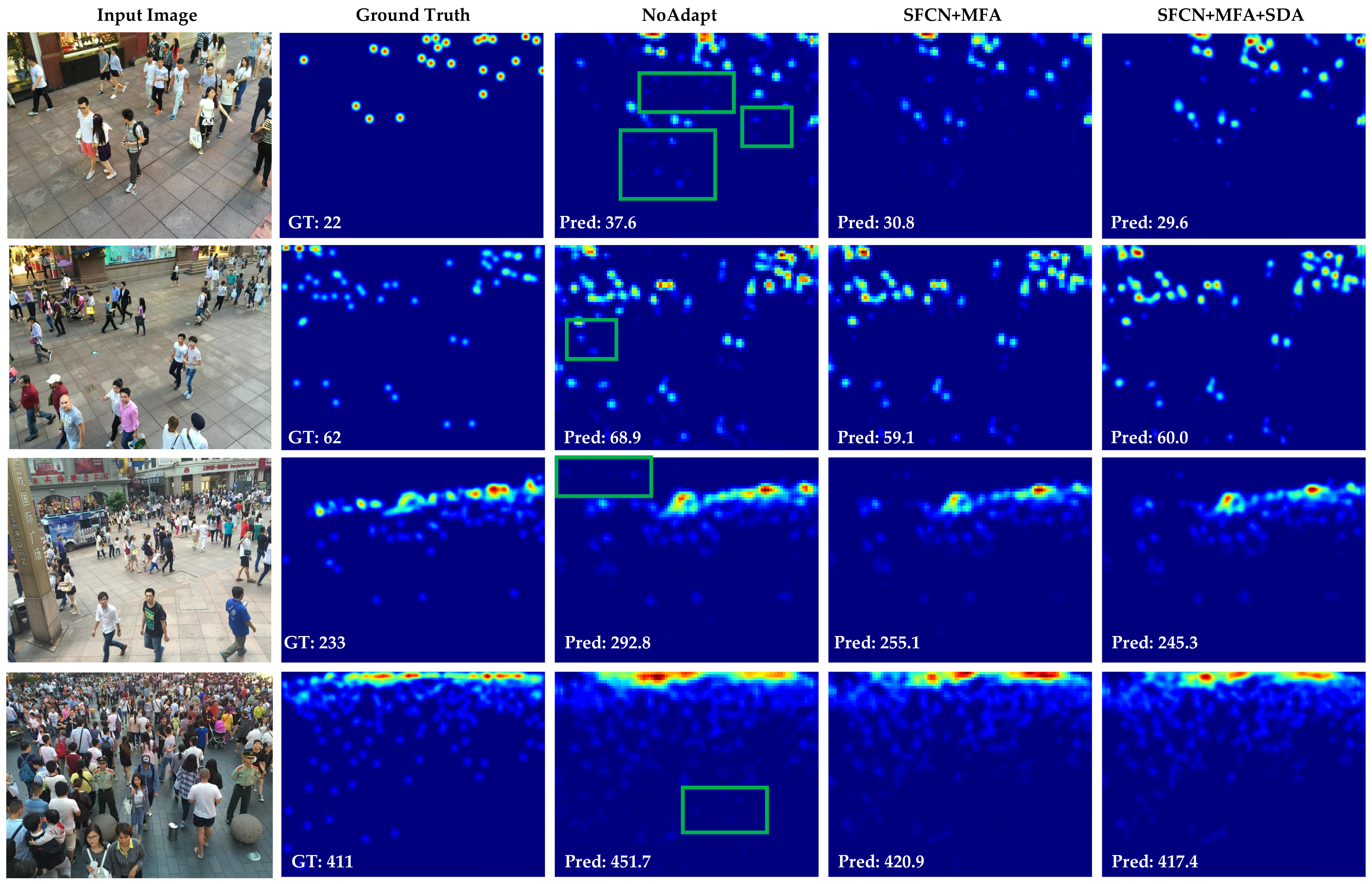}
	\caption{Exemplar results of adaptation from GCC to Shanghai Tech Part B dataset. From left to right: input image, ground truth, and the predictions of NoAdpt, SFCN+MFA, and SFCN+MFA+SDA.}\label{re-examplars}
\end{figure*}

Take the results of Shanghai Tech B as an example, MFA decreases the {\color{red}{22.3\%}} MAE and the {\color{red}{12.7\%}} MSE, which implies the proposed MFA can effectively reduce the domain gap between synthetic and real-world data. When introducing SDA, the errors are further alleviated. As a result, our proposed method achieves the MAE of 16.4 and the MSE of 25.4,  which is better than the result of SE Cycle GAN (MAE/MSE: 19.9/28.3). 

Fig. \ref{re-examplars} shows the visualization results on Shanghai Tech Part B dataset. From Column 3, NoAdapt produces some estimation errors on the background, especially in the green boxes. When adopting MFA and SDA, the aforementioned errors are effectively reduced. By comparing Column 4 and 5, we find that SDA yields more reasonable and finer density maps (recommend readers to zoom in the image for comparison). As a result, the counting values of Column 5 are closer to the ground truth than that of Column 4.

\subsection{Ablation Study on Shanghai Tech B}
\label{quality}

\textbf{Analysis of different modules.} Table \ref{Table-quality} reports the results of our proposed step-wise methods: No Adaptation (NoAdpt), SFCN+MFA, SFCN+SDA and SFCN+MFA+SDA. The error reductions of SDA are smaller than those of MFA. Compared with NoAdpt, MFA decreases {\color{red}{22.3\%}} in MAE and  {\color{red}{12.7\%}} in MSE, but SDA only reduces {\color{red}{11.7/5.0\%}} in terms of MAE/MSE. The main reason is that SDA mainly aims to improve the quality of maps, but MFA focuses on aiding SFCN to extract domain-invariant features, which is the core of reducing the domain gap.

\textbf{Comparison with Cycle-GAN-style methods.} Here, we compare the quality of density maps of our proposed algorithms with Cycle-GAN-style methods \mbox{\cite{zhu2017unpaired,hoffman2017cycada,wang2019learning}} on Shanghai Tech B in Table \mbox{\ref{Table-quality}}. All methods adopt the same SFCN with VGG-16 backbone as the counter network. As for the four metrics, the proposed adaptation algorithms perform better than Cycle-GAN-style methods. In terms of the methodology, GAN and SFCN are trained separately in Cycle-GAN-style methods: 1) train GAN and translate the image; 2) train SFCN using translated images. This separated training scheme can not attain a good result. In our method, the training of SFCN and adversarial networks is applied iteratively, which guarantees the entire models achieve better performance.

\textbf{Comparison with feature-level adversarial learning methods.} Compared with FCN Wld w/o CA \mbox{\cite{hoffman2016fcns}} and CODA \mbox{\cite{8784811}} that also attempt to extract domain-invariant features in a specific layer of the networks, the proposed MFA outperforms them (MAE: 17.3 (MFA) \emph{vs.} 23.1 \mbox{\cite{hoffman2016fcns}} and 19.2 \mbox{\cite{8784811}}) due to it conducts adversarial learning on the high-level semantic features and multiple layers, which causes that the model extracts more effective domain-invariant features. In the experiment, note that FCN Wld contains category adaptation (CA) for pixel-wise classification. Thus, we remove CA from FCN Wld. At the same time, we replace the classification layer with a regression layer in FCN Wld. Correspondingly, the task loss is MSE loss instead of Cross Entropy Loss. 

\begin{table}[htbp]
	\small
	\centering
	\caption{The performance on Shanghai Tech Part B.}
	\setlength{\tabcolsep}{1.2mm}{
	\begin{tabular}{c|c|cc|cc}
		\whline
		Methods 	&Src data			 &MAE 		&MSE	 & PSNR &SSIM		\\
		\whline
		NoAdpt \cite{wang2019learning}     &GCC   &22.8 &30.6 &24.66 &0.715	\\
		FCN Wld w/o CA \cite{hoffman2016fcns} &GCC  &23.1 &29.4 &24.12  &0.732	\\
		Cycle GAN  \cite{zhu2017unpaired} &GCC  &25.4 &39.7 & 24.60 &0.763	\\
		CyCADA \cite{hoffman2017cycada} &GCC &{18.7} &{26.6} &{25.58}  &{0.826}	\\
		SE Cycle GAN \cite{wang2019learning} &GCC   &19.9 &28.3 &24.78 & 0.765	\\
		CODA \cite{8784811} &GCC   &19.2 &28.5 &24.63 & 0.803	\\
		
		NoAdpt (ours) &GCC &22.3 &29.9 &25.02 &0.811 \\
		SFCN+MFA &GCC  &17.3 &26.1&25.37 &0.837 \\
		SFCN+SDA &GCC  &19.7 &28.4&25.11 &0.832 \\
		SFCN+MFA+MD+SPR &GCC  &16.2 &24.9 &25.45 &0.846\\
		SFCN+MFA+SDA&GCC &{\underline{16.0}} &{\underline{24.7}}& {\underline{25.62}}& \textbf{\underline{0.856}}\\
		\hline
		\hline
		NoAdpt \cite{8784811} &SHT A   &27.3 &36.2 &- &- \\		
		CODA \cite{8784811} &SHT A   &15.9 &26.9 & - & - \\
		SFCN+MFA+SDA&SHT A &\textbf{\underline{14.8}} &\textbf{\underline{21.9}}& \textbf{\underline{25.89}}& {\underline{0.844}}\\
		\whline	
	\end{tabular}
}\label{Table-quality}
\end{table}

\begin{figure}[t]
	\centering
	\includegraphics[width=.48\textwidth]{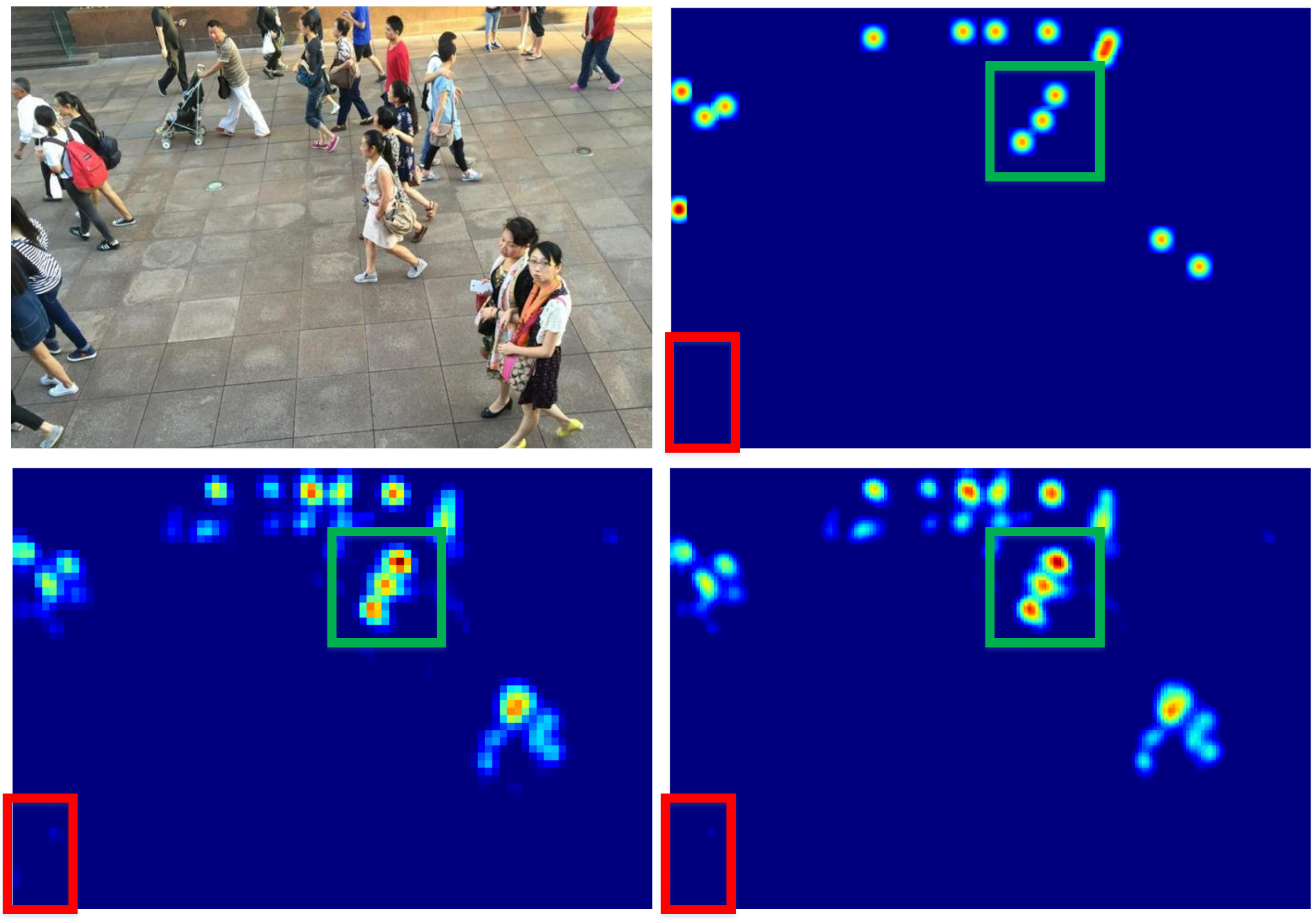}
	\caption{The comparison of with/without MR. Row 1: original image, ground truth; Row 2: without MR, with MR.}\label{com-MR}
\end{figure}

\textbf{Effect of Map Refiner.} In our proposed method, Map Refiner (MR) is an independent network, which is trained on the testing set on GCC dataset. It can effectively refine the quality of SFCN's prediction maps. Table \ref{Table-MR} lists the quantitative results with/without MR. From it, we indeed find MR prompts the performance of counting and density map quality. In order to intuitively show the effect of MR, the density maps are compared in Fig. \ref{com-MR}. According to the comparison, there are three main effects of MR: 1) MR produces finer density maps for the entire outputs; 2) MR yields more independent density region, such as the green box; 3) MR reduces the estimation errors for low-response regions, such as the red box. In general, the proposed MR outputs the more reasonable and finer density maps.

\begin{table}[htbp]
	\centering
	\caption{The effect of each module in SDA on ShanghaiTech B.}
	\setlength{\tabcolsep}{1.0mm}{
		\begin{tabular}{c|cc|cc}
			\whline
			Methods 				 &MAE 		&MSE	 &PSNR &SSIM		\\
			\whline
			SFCN+MFA+MD   &16.5 &25.8 &25.41 &0.845\\
			SFCN+MFA+MD+SPR   &16.2 &24.9 &25.45 &0.846\\
			SFCN+MFA+MD+SPR+MR &\textbf{\underline{16.0}} &\textbf{\underline{24.7}}& \textbf{\underline{25.62}}& \textbf{\underline{0.856}}\\
			\whline	
		\end{tabular}
	}\label{Table-MR}
\end{table}

\section{Analysis and Discussion}

\subsection{Selection of Feature Maps in MFA}
In MFA, the feature maps of Dilatation and Spatial Module are extracted to train the two domain discriminators. In fact, there are many potential feature maps in SFCN that are selected as the inputs for domain discriminators. In this section, we analyze the effects under different combinations of different layers' outputs on experimental results. To be specific, three types of feature maps are selected, namely the outputs of conv4\_3, Dilatation Module and Spatial Module. 

\begin{table}[htbp]
	\centering	
	\caption{The comparison results of different combinations of feature maps. }
	\setlength{\tabcolsep}{1.5mm}{
		\begin{tabular}{cIc|c|cIc|c}
			\whline
			\multirow{2}{*}{Methods} &\multicolumn{3}{cI}{Combination} 	 &\multicolumn{2}{c}{Errors}			\\
			\cline{2-6} 
			&conv4\_3 &Dilation &Spatial &MAE & MSE		\\
			\whline
			NoAdapt&    &  &  & 22.3& 29.9    \\	
			\whline
			\multirow{3}{*}{single}&\rmark   &        &       &20.3 & 28.1	\\
			\cline{2-6} 
			&		 &\rmark  &       &18.5 & \underline{27.2} 	\\
			\cline{2-6} 
			&		 &        &\rmark & \underline{18.4} & 29.1 	\\
			\whline
			\multirow{3}{*}{double}& \rmark   &\rmark  &        &18.6&27.8	\\
			\cline{2-6} 
			&\rmark   &        &\rmark	  & 19.3 &29.7  \\
			\cline{2-6} 
			&         &\rmark  &\rmark  & \textbf{\underline{17.3}} &\textbf{\underline{26.1}}     \\		
			\whline
			triple & \rmark   &\rmark  &\rmark   &\textbf{17.3} &26.3     \\	
			\whline	
		\end{tabular}
	}\label{combinations}
\end{table}

Table \ref{combinations} reports the estimation errors under different settings on Shanghai Tech Part B. From the results of single-level adversarial experiments, the domain gap can be reduced more effectively on the deep layer (after Dilation and Spatial) than the shallow layer (conv4\_3). In the double-level experiments, the combination of Dilation and Spatial Module is the best, which achieves the MAE of 17.3 and the MSE of 26.1. When adopting triple-level adversarial learning, the results are very close to that of Dilation+Spatial. We think the deep-layer adversarial training can reduce the domain gap to the maximum extent. It is not necessary that introducing shallow-layer adversarial learning. Therefore,  Dilation+Spatial is the final combination in MFA. 

Furthermore, we discuss why Dilation+Spatial is the best combination. The main reasons are: 1) the two modules encode more large-range contextual information than that of the backbone. 2) crowd density is a type of high-level features. For a deep CNN, shallow layers only extract low-level features, such as texture, edge or other appearance characteristics. Deep layers can extract high-level semantic features. In the two data domains, difference of low-level features of is an objective reality. But the crowd density features are the same. Thus, we select Dilation and Spatial Module to construct adversarial learning, which makes models to learn domain-invariant features.

\subsection{Adaptation between Real-world Datasets}

The domain gap does not only exist in the adaptation between the synthetic and the real world but also in the transferring process of different real scenes. For example, real-world datasets are very different in terms of scene attributes, cameras, and so on. In this section, we select two typical real-world datasets (Mall and UCSD) to evaluate the proposed method. These two datasets are very different: Mall is an indoor crowd dataset captured by RGB cameras but UCSD is an outdoor crowd dataset captured by gray-scale sensors. Table \ref{Table-realDA} reports the results of two different adaptation experiments (Mall$\rightarrow$UCSD and UCSD$\rightarrow$Mall). From it,  the proposed method effectively reduces the domain gap between different datasets and attains an acceptable counting result, which shows that our method is important for landing the counter network in real life.

\begin{table}[htbp]
	\centering	
	\caption{The performance of adaptation between the two real-world datasets.}
	\begin{tabular}{c|cc|cc}
		\whline
		\multirow{2}{*}{Methods} &\multicolumn{2}{c|}{Mall$\rightarrow$UCSD} 	 &\multicolumn{2}{c}{UCSD$\rightarrow$Mall} 		\\
		\cline{2-5} 
		&MAE &MSE &MAE &MSE		\\
		\whline
		NoAdpt   &15.80 &16.14 &3.01 &3.74\\
		\hline
		CODA \cite{8784811}   &- &- &3.38 &4.15\\
		\hline
		SFCN+MFA+SDA &\textbf{2.08} &\textbf{2.58} &\textbf{2.66}& \textbf{3.32}\\
		\whline	
	\end{tabular}\label{Table-realDA}
\end{table}

\subsection{Generalization Ability on Other Models}

In this paper, we conduct the adaptation experiments on the SFCN. In fact, the proposed method can be applied to any FCN-based crowd counter. Here, we adopt two classical crowd counters, MCNN \cite{zhang2016single} and CSRNet \cite{li2018csrnet},  to verify the proposed adaption method on Shanghai Tech Part B Dataset. For CSRNet, we select the feature maps of the VGG-16 backbone and dilated convolutional layers as the input for Multi-level Feature-aware Adaptation (MFA). Since MCNN only contains single-stage convolution, we have to apply Single-level Feature-aware Adaptation (SFA) on MCNN.

\begin{table}[htbp]
	\centering	
	\caption{The comparison of different crowd counter in the proposed adaptation method.}
	\setlength{\tabcolsep}{2.0mm}{
		\begin{tabular}{c|cc|cc}
			\whline
			\multirow{2}{*}{Methods} &\multicolumn{2}{c|}{NoAdpt} 	 &\multicolumn{2}{c}{Adapted} 		\\
			\cline{2-5} 
			&MAE &MSE &MAE &MSE		\\
			\whline
			MCNN  &66.1 &82.8 &\underline{28.0} &\underline{43.9}\\
			\hline
			CSRNet &23.0 &33.9 &\underline{18.1}& \underline{26.9}\\
			\hline
			SFCN &\textbf{22.3} &\textbf{29.9} &\textbf{\underline{16.4}} & \textbf{\underline{25.4}}\\
			\whline	
		\end{tabular}
	}\label{Table-gen}
\end{table}

Table \ref{Table-gen} lists the comparison results on the three different crowd counters. From it, we find the estimation errors are significantly decreased after adaptation. This phenomenon indicates our method can be generalized to other CNN-based crowd counters to reduce the domain gap.

\subsection{Adaptation to Inconsistent Real Data}

Usually, the same surveillance cameras are equipped in a city or a specific region by the government. As a result, the captured scenes are highly consistent. Considering that, this work focuses on the domain adaptation to the consistent crowd scenes, which is practical in the real scene application. For the traditional crowd counting, there are two real-world datasets that contains inconsistent crowd scenes collected from Internet, namely, Shanghai Tech Part A \cite{zhang2016single} and UCF-QNRF \cite{idrees2018composition}, of which images are captured by people instead of surveillance cameras. Here, we verify the proposed method on these datasets.

\begin{table}[htbp]
	\centering
	
	\caption{The performance of No Adaptation (NoAdapt), Cycle GAN\cite{zhu2017unpaired}, SE Cycle GAN \cite{wang2019learning} and our approaches on Shanghai Tech Part A and UCF-QNRF dataset.}
	\setlength{\tabcolsep}{1.5mm}{
		\begin{tabular}{c|cIc|cIc|c}
			\whline
			\multirow{2}{*}{Method}	&\multirow{2}{*}{DA} &\multicolumn{2}{cI}{SHT A} & \multicolumn{2}{c}{UCF-QNRF} \\
			\cline{3-6} 
			& & MAE &MSE & MAE &MSE  \\
			\whline
			NoAdpt \cite{wang2019learning}  &\xmark &160.0 &216.5 &275.5 &458.5 \\
			\hline
			Cycle GAN \cite{zhu2017unpaired} &\rmark &143.3 &204.3 &257.3 &400.6  \\
			\hline	
			SE Cycle GAN \cite{wang2019learning}  &\rmark&\textbf{123.4} &\textbf{193.4} &\textbf{230.4} &\textbf{384.5} \\
			\whline
			NoAdpt (ours)  &\xmark &156.4 &210.7 &269.5 &480.2 \\
			\hline	
			SFCN+MFA+SDA &\rmark &144.6 &200.6 &255.4 &407.9   \\
			\whline			
		\end{tabular}
	}\label{Table-a-ucf}
\end{table}

\begin{figure*}[h]
	\centering
	\includegraphics[width=.98\textwidth]{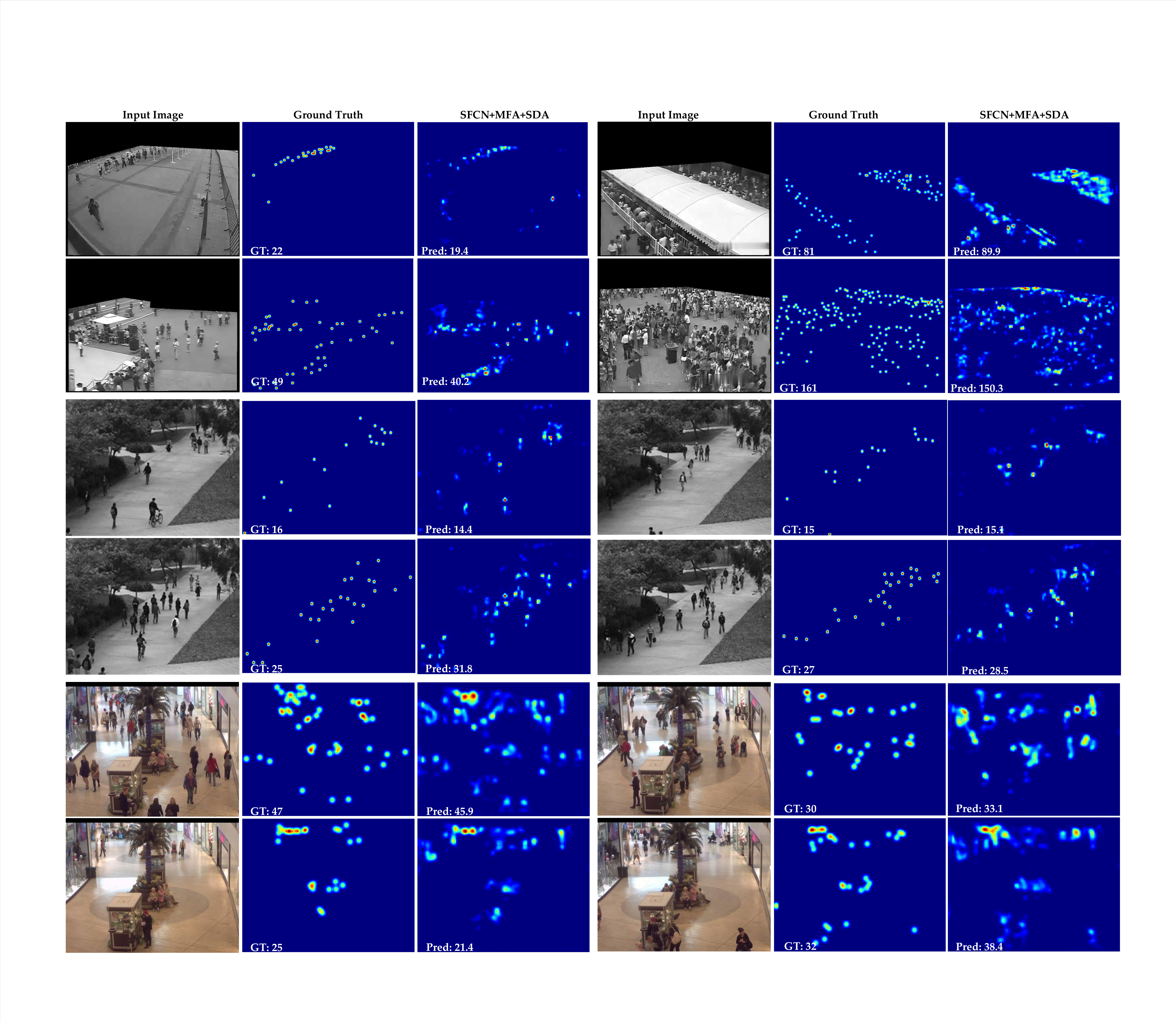}
	\caption{Exemplar results of adaptation from GCC to WorldExpo'10, UCSD, and Mall dataset.}\label{mix}
\end{figure*}

The results on these two datasets are shown in Table \ref{Table-a-ucf}. From it, we find the results of the proposed method is far from that of SE Cycle GAN \cite{wang2019learning}. The main reason is that inconsistent real data causes that the GANs ($\boldsymbol{G}$ and $\boldsymbol{D_i}$ ($i = 1,2,3$)) are hard to converge. Especially, the loss for $\boldsymbol{G}$ suffers from severe fluctuations. 

In summary, our method is more suitable for adaptation from synthetic data to real-world surveillance scenes.

\subsection{The Effect of Large Kernel in MR}

In Section \mbox{\ref{mr}}, we design a Map Refiner with large kernel size to cover a large respective field. Here, we conduct some experiments to verify our network design. To be specific, three Map Refiners are designed with different settings for the number of kernel sizes: $[13,9,5,9,13]$, $[9,5,3,5,9]$ and $[7,5,3,5,7]$ (Fig. \mbox{\ref{Fig-MRnet}}'s setting is the first one). Table \mbox{\ref{Table-mr-design}} reports the refiner results in terms of MAE, MSE, PSNR and SSIM. From it, we find the setting of $[13,9,5,9,13]$ obtains the best refinement performance. The main reason is that large kernel cover the large-range spatial information to reduce the low-response noises and improve the reconstruction performance for a whole predicted head density region. 

\begin{table}[htbp]
	\centering	
	\caption{The comparison of different kernel sizes in the Map Refiner on ShanghaiTech Part B.}
	\setlength{\tabcolsep}{2.0mm}{
		\begin{tabular}{c|cc|cc}
			\whline
			Settings &MAE &MSE &PSNR &SSIM	 		\\
			\whline
			None  &16.24 &24.91 &25.45 & 0.846\\
			\hline
			$[13,9,5,9,13]$ &16.02 &24.73 &25.62 & 0.856\\
			\hline
			$[9,5,3,5,9]$ &16.09 &24.81 &25.58 & 0.849\\
			\hline
			$[7,5,3,5,7]$ &16.20 &24.87 &25.47 & 0.841\\
			\whline	
		\end{tabular}
	}\label{Table-mr-design}
\end{table}

\subsection{Introducing SPR into Classical Supervised Models}

In this section, we implement SPR in the training of MCNN and CSRNet on ShanghaiTech Part B dataset. Table \mbox{\ref{Table-spr}} illustrates the improvements after introducing the SPR in the Traditional Supervision (TS). From the table, we find the estimation errors are reduced significantly. Notably, take MCNN as an example, MAE/MSE decreases from 23.3/40.2 to 21.8/37.7 (relatively 6.4/6.2\%) on ShanghaiTech Part B. In fact, SPR can be treated as a combination of multi-scale training and self-supervised algorithms, which makes the model more robust.

\begin{table}[htbp]
	\centering	
	\caption{The effect of SPR in traditional supervised models on ShanghaiTech Part B.}
	\setlength{\tabcolsep}{2.0mm}{
		\begin{tabular}{c|c|c}
			\whline
			Method &TS 	& TS+SPR 	\\
			\whline
			MCNN  &23.3/40.2 &21.8/37.7 \\
			\hline
			CSRNet &10.6/17.1 &10.1/15.4\\
			\whline	
		\end{tabular}
	}\label{Table-spr}
\end{table}

\subsection{Visualization Results}

This section demonstrates the visualization results on WorldExpo'10 \mbox{\cite{zhang2016data}}, Mall \mbox{\cite{chen2012feature}} and UCSD \mbox{\cite{chan2008privacy}} datasets in Fig. \mbox{\ref{mix}}. In general, the adaptation results are able to reflect the crowd density and predict the number of people approximately. However, compared with ground truth, there are many error estimations in the background region, especially in some places far away from the camera. The main reason is these places are very similar to head in terms of texture. Note that there are some shifts in the visualization results on the UCSD dataset. The main reason is that the key-point location in UCSD is the center of the person, but the key-point location in GCC is the center of the head.

\section{Conclusion}

In this work, we present a GAN-based adaptation method for crowd counting by learning from synthetic data and the corresponding free labels. The proposed method consists of two modules: Multi-level Feature-aware Adaptation (MFA) and a Structured Density map Alignment (SDA). The former module focuses on reducing the domain gap between synthetic and real data at the feature level, which is the first attempt in crowd counting. The latter aims to produce reasonable and fine density maps on the unseen data. Experimental results show that the proposed method outperforms the state-of-the-art performance for the same task. For the future work, since the high-level semantic information (such as the structured features of persons or groups) are more robust and invariant than pixel-level features in the cross-domain problem, we will attempt to introduce these features into domain adaptation to prompt the counting performance in the real world.

\bibliographystyle{IEEEtran}
\bibliography{IEEEabrv,reference}

\begin{thebibliography}{10}
\providecommand{\url}[1]{#1}
\csname url@samestyle\endcsname
\providecommand{\newblock}{\relax}
\providecommand{\bibinfo}[2]{#2}
\providecommand{\BIBentrySTDinterwordspacing}{\spaceskip=0pt\relax}
\providecommand{\BIBentryALTinterwordstretchfactor}{4}
\providecommand{\BIBentryALTinterwordspacing}{\spaceskip=\fontdimen2\font plus
\BIBentryALTinterwordstretchfactor\fontdimen3\font minus
  \fontdimen4\font\relax}
\providecommand{\BIBforeignlanguage}[2]{{%
\expandafter\ifx\csname l@#1\endcsname\relax
\typeout{** WARNING: IEEEtran.bst: No hyphenation pattern has been}%
\typeout{** loaded for the language `#1'. Using the pattern for}%
\typeout{** the default language instead.}%
\else
\language=\csname l@#1\endcsname
\fi
#2}}
\providecommand{\BIBdecl}{\relax}
\BIBdecl

\bibitem{onoro2016towards}
D.~Onoro-Rubio and R.~J. L{\'o}pez-Sastre, ``Towards perspective-free object
  counting with deep learning,'' in \emph{European Conference on Computer
  Vision}.\hskip 1em plus 0.5em minus 0.4em\relax Springer, 2016, pp. 615--629.

\bibitem{hu2016dense}
Y.~Hu, H.~Chang, F.~Nian, Y.~Wang, and T.~Li, ``Dense crowd counting from still
  images with convolutional neural networks,'' \emph{Journal of Visual
  Communication and Image Representation}, vol.~38, pp. 530--539, 2016.

\bibitem{babu2017switching}
D.~Babu~Sam, S.~Surya, and R.~Venkatesh~Babu, ``Switching convolutional neural
  network for crowd counting,'' in \emph{Proceedings of the IEEE Conference on
  Computer Vision and Pattern Recognition}, 2017, pp. 5744--5752.

\bibitem{babu2018divide}
D.~Babu~Sam, N.~N. Sajjan, R.~Venkatesh~Babu, and M.~Srinivasan, ``Divide and
  grow: capturing huge diversity in crowd images with incrementally growing
  cnn,'' in \emph{Proceedings of the IEEE Conference on Computer Vision and
  Pattern Recognition}, 2018, pp. 3618--3626.

\bibitem{shi2019revisiting}
M.~Shi, Z.~Yang, C.~Xu, and Q.~Chen, ``Revisiting perspective information for
  efficient crowd counting,'' in \emph{Proceedings of the IEEE Conference on
  Computer Vision and Pattern Recognition}, 2019, pp. 7279--7288.

\bibitem{wan2019residual}
J.~Wan, W.~Luo, B.~Wu, A.~B. Chan, and W.~Liu, ``Residual regression with
  semantic prior for crowd counting,'' in \emph{Proceedings of the IEEE
  Conference on Computer Vision and Pattern Recognition}, 2019, pp. 4036--4045.

\bibitem{liu2019crowd}
L.~Liu, Z.~Qiu, G.~Li, S.~Liu, W.~Ouyang, and L.~Lin, ``Crowd counting with
  deep structured scale integration network,'' \emph{arXiv preprint
  arXiv:1908.08692}, 2019.

\bibitem{8949751}
Y.~{Zhou}, J.~{Yang}, H.~{Li}, T.~{Cao}, and S.~{Kung}, ``Adversarial learning
  for multiscale crowd counting under complex scenes,'' \emph{IEEE Transactions
  on Cybernetics}, pp. 1--10, 2020.

\bibitem{chan2008privacy}
A.~B. Chan, Z.-S.~J. Liang, and N.~Vasconcelos, ``Privacy preserving crowd
  monitoring: Counting people without people models or tracking,'' in
  \emph{Computer Vision and Pattern Recognition, 2008. CVPR 2008. IEEE
  Conference on}.\hskip 1em plus 0.5em minus 0.4em\relax IEEE, 2008, pp. 1--7.

\bibitem{chen2012feature}
K.~Chen, C.~C. Loy, S.~Gong, and T.~Xiang, ``Feature mining for localised crowd
  counting.'' in \emph{BMVC}, vol.~1, no.~2, 2012, p.~3.

\bibitem{zhang2016single}
Y.~Zhang, D.~Zhou, S.~Chen, S.~Gao, and Y.~Ma, ``Single-image crowd counting
  via multi-column convolutional neural network,'' in \emph{Proceedings of the
  IEEE conference on computer vision and pattern recognition}, 2016, pp.
  589--597.

\bibitem{zhang2016data}
C.~Zhang, K.~Kang, H.~Li, X.~Wang, R.~Xie, and X.~Yang, ``Data-driven crowd
  understanding: a baseline for a large-scale crowd dataset,'' \emph{IEEE
  Transactions on Multimedia}, vol.~18, no.~6, pp. 1048--1061, 2016.

\bibitem{jiang2019learning}
X.~Jiang, L.~Zhang, P.~Lv, Y.~Guo, R.~Zhu, Y.~Li, Y.~Pang, X.~Li, B.~Zhou, and
  M.~Xu, ``Learning multi-level density maps for crowd counting,'' \emph{IEEE
  transactions on neural networks and learning systems}, 2019.

\bibitem{7434629}
V.~J. {Kok} and C.~S. {Chan}, ``Grcs: Granular computing-based crowd
  segmentation,'' \emph{IEEE Transactions on Cybernetics}, vol.~47, no.~5, pp.
  1157--1168, May 2017.

\bibitem{7165622}
A.~S. {Rao}, J.~{Gubbi}, S.~{Marusic}, and M.~{Palaniswami}, ``Crowd event
  detection on optical flow manifolds,'' \emph{IEEE Transactions on
  Cybernetics}, vol.~46, no.~7, pp. 1524--1537, 2016.

\bibitem{li2014crowded}
T.~Li, H.~Chang, M.~Wang, B.~Ni, R.~Hong, and S.~Yan, ``Crowded scene analysis:
  A survey,'' \emph{IEEE transactions on circuits and systems for video
  technology}, vol.~25, no.~3, pp. 367--386, 2014.

\bibitem{yuan2018structured}
Y.~Yuan, Y.~Feng, and X.~Lu, ``Structured dictionary learning for abnormal
  event detection in crowded scenes,'' \emph{Pattern Recognition}, vol.~73, pp.
  99--110, 2018.

\bibitem{8802265}
K.~K. {Santhosh}, D.~P. {Dogra}, P.~P. {Roy}, and B.~B. {Chaudhuri},
  ``Trajectory-based scene understanding using dirichlet process mixture
  model,'' \emph{IEEE Transactions on Cybernetics}, pp. 1--14, 2019.

\bibitem{zhao2019property}
B.~Zhao, X.~Li, and X.~Lu, ``Property-constrained dual learning for video
  summarization,'' \emph{IEEE Transactions on Neural Networks and Learning
  Systems}, 2019.

\bibitem{DBLP:conf/cvpr/Xiong0G020}
Z.~Xiong, Y.~Yuan, N.~Guo, and Q.~Wang, ``Variational context-deformable
  convnets for indoor scene parsing,'' in \emph{2020 {IEEE/CVF} Conference on
  Computer Vision and Pattern Recognition, {CVPR} 2020, Seattle, WA, USA, June
  13-19, 2020}.\hskip 1em plus 0.5em minus 0.4em\relax {IEEE}, 2020, pp.
  3991--4001.

\bibitem{yuan2017tracking}
Y.~Yuan, Y.~Lu, and Q.~Wang, ``Tracking as a whole: Multi-target tracking by
  modeling group behavior with sequential detection,'' \emph{IEEE Transactions
  on Intelligent Transportation Systems}, vol.~18, no.~12, pp. 3339--3349,
  2017.

\bibitem{6786497}
Y.~{Han}, Y.~{Yang}, Y.~{Yan}, Z.~{Ma}, N.~{Sebe}, and X.~{Zhou},
  ``Semisupervised feature selection via spline regression for video semantic
  recognition,'' \emph{IEEE Transactions on Neural Networks and Learning
  Systems}, vol.~26, no.~2, pp. 252--264, 2015.

\bibitem{zhao2019cam}
B.~Zhao, X.~Li, and X.~Lu, ``Cam-rnn: Co-attention model based rnn for video
  captioning,'' \emph{IEEE Transactions on Image Processing}, vol.~28, no.~11,
  pp. 5552--5565, 2019.

\bibitem{DBLP:journals/ijon/XiongYW20}
Z.~Xiong, Y.~Yuan, and Q.~Wang, ``{MSN:} modality separation networks for
  {RGB-D} scene recognition,'' \emph{Neurocomputing}, vol. 373, pp. 81--89,
  2020.

\bibitem{deng2009imagenet}
J.~Deng, W.~Dong, R.~Socher, L.-J. Li, K.~Li, and L.~Fei-Fei, ``Imagenet: A
  large-scale hierarchical image database,'' in \emph{2009 IEEE conference on
  computer vision and pattern recognition}, 2009, pp. 248--255.

\bibitem{zhao2019weather}
B.~Zhao, L.~Hua, X.~Li, X.~Lu, and Z.~Wang, ``Weather recognition via
  classification labels and weather-cue maps,'' \emph{Pattern Recognition},
  vol.~95, pp. 272--284, 2019.

\bibitem{redmon2016you}
J.~Redmon, S.~Divvala, R.~Girshick, and A.~Farhadi, ``You only look once:
  Unified, real-time object detection,'' in \emph{Proceedings of the IEEE
  conference on computer vision and pattern recognition}, 2016, pp. 779--788.

\bibitem{ijcai2020-416}
Z.~Wang, F.~Nie, L.~Tian, R.~Wang, and X.~Li, ``Discriminative feature
  selection via a structured sparse subspace learning module,'' in
  \emph{Proceedings of the Twenty-Ninth International Joint Conference on
  Artificial Intelligence, {IJCAI-20}}, 2020, pp. 3009--3015.

\bibitem{wang2019learning}
Q.~Wang, J.~Gao, W.~Lin, and Y.~Yuan, ``Learning from synthetic data for crowd
  counting in the wild,'' \emph{arXiv preprint arXiv:1903.03303}, 2019.

\bibitem{liu2018leveraging}
X.~Liu, J.~van~de Weijer, and A.~D. Bagdanov, ``Leveraging unlabeled data for
  crowd counting by learning to rank,'' \emph{arXiv preprint arXiv:1803.03095},
  2018.

\bibitem{sam2019unsupervised}
D.~B. Sam, N.~N. Sajjan, H.~Maurya, and R.~V. Babu, ``Almost unsupervised
  learning for dense crowd counting,'' 2019.

\bibitem{wang2020pixel}
Q.~Wang, J.~Gao, W.~Lin, and Y.~Yuan, ``Pixel-wise crowd understanding via
  synthetic data,'' \emph{International Journal of Computer Vision}, pp. 1--21,
  2020.

\bibitem{kong2006viewpoint}
D.~Kong, D.~Gray, and H.~Tao, ``A viewpoint invariant approach for crowd
  counting,'' in \emph{18th International Conference on Pattern Recognition
  (ICPR'06)}, vol.~3.\hskip 1em plus 0.5em minus 0.4em\relax IEEE, 2006, pp.
  1187--1190.

\bibitem{rabaud2006counting}
V.~Rabaud and S.~Belongie, ``Counting crowded moving objects,'' in \emph{2006
  IEEE Computer Society Conference on Computer Vision and Pattern Recognition
  (CVPR'06)}, vol.~1.\hskip 1em plus 0.5em minus 0.4em\relax IEEE, 2006, pp.
  705--711.

\bibitem{sindagi2017generating}
V.~A. Sindagi and V.~M. Patel, ``Generating high-quality crowd density maps
  using contextual pyramid cnns,'' in \emph{2017 IEEE International Conference
  on Computer Vision (ICCV)}.\hskip 1em plus 0.5em minus 0.4em\relax IEEE,
  2017, pp. 1879--1888.

\bibitem{sam2018top}
D.~B. Sam and R.~V. Babu, ``Top-down feedback for crowd counting convolutional
  neural network,'' in \emph{Thirty-Second AAAI Conference on Artificial
  Intelligence}, 2018.

\bibitem{li2018csrnet}
Y.~Li, X.~Zhang, and D.~Chen, ``Csrnet: Dilated convolutional neural networks
  for understanding the highly congested scenes,'' in \emph{Proceedings of the
  IEEE Conference on Computer Vision and Pattern Recognition}, 2018, pp.
  1091--1100.

\bibitem{cao2018scale}
X.~Cao, Z.~Wang, Y.~Zhao, and F.~Su, ``Scale aggregation network for accurate
  and efficient crowd counting,'' in \emph{Proceedings of the European
  Conference on Computer Vision (ECCV)}, 2018, pp. 734--750.

\bibitem{ranjan2018iterative}
V.~Ranjan, H.~Le, and M.~Hoai, ``Iterative crowd counting,'' in
  \emph{Proceedings of the European Conference on Computer Vision (ECCV)},
  2018, pp. 270--285.

\bibitem{gao2019scar}
J.~Gao, Q.~Wang, and Y.~Yuan, ``Scar: Spatial-/channel-wise attention
  regression networks for crowd counting,'' \emph{Neurocomputing}, vol. 363,
  pp. 1--8, 2019.

\bibitem{idrees2018composition}
H.~Idrees, M.~Tayyab, K.~Athrey, D.~Zhang, S.~Al-Maadeed, N.~Rajpoot, and
  M.~Shah, ``Composition loss for counting, density map estimation and
  localization in dense crowds,'' \emph{arXiv preprint arXiv:1808.01050}, 2018.

\bibitem{liu2019context}
W.~Liu, M.~Salzmann, and P.~Fua, ``Context-aware crowd counting,'' in
  \emph{Proceedings of the IEEE Conference on Computer Vision and Pattern
  Recognition}, 2019, pp. 5099--5108.

\bibitem{sindagi2019multi}
V.~A. Sindagi and V.~M. Patel, ``Multi-level bottom-top and top-bottom feature
  fusion for crowd counting,'' \emph{arXiv preprint arXiv:1908.10937}, 2019.

\bibitem{jiang2019crowd}
X.~Jiang, Z.~Xiao, B.~Zhang, X.~Zhen, X.~Cao, D.~Doermann, and L.~Shao, ``Crowd
  counting and density estimation by trellis encoder-decoder networks,'' in
  \emph{Proceedings of the IEEE Conference on Computer Vision and Pattern
  Recognition}, 2019, pp. 6133--6142.

\bibitem{sindagi2017cnn}
V.~A. Sindagi and V.~M. Patel, ``Cnn-based cascaded multi-task learning of
  high-level prior and density estimation for crowd counting,'' in \emph{2017
  14th IEEE International Conference on Advanced Video and Signal Based
  Surveillance (AVSS)}.\hskip 1em plus 0.5em minus 0.4em\relax IEEE, 2017, pp.
  1--6.

\bibitem{gao2019pcc}
J.~Gao, Q.~Wang, and X.~Li, ``Pcc net: Perspective crowd counting via spatial
  convolutional network,'' \emph{IEEE Transactions on Circuits and Systems for
  Video Technology}, 2019.

\bibitem{zhao2019leveraging}
M.~Zhao, J.~Zhang, C.~Zhang, and W.~Zhang, ``Leveraging heterogeneous auxiliary
  tasks to assist crowd counting,'' in \emph{Proceedings of the IEEE Conference
  on Computer Vision and Pattern Recognition}, 2019, pp. 12\,736--12\,745.

\bibitem{he2019dynamic}
G.~He, Z.~Ma, B.~Huang, B.~Sheng, and Y.~Yuan, ``Dynamic region division for
  adaptive learning pedestrian counting,'' in \emph{2019 IEEE International
  Conference on Multimedia and Expo (ICME)}.\hskip 1em plus 0.5em minus
  0.4em\relax IEEE, 2019, pp. 1120--1125.

\bibitem{Lian_2019_CVPR}
D.~Lian, J.~Li, J.~Zheng, W.~Luo, and S.~Gao, ``Density map regression guided
  detection network for rgb-d crowd counting and localization,'' in \emph{The
  IEEE Conference on Computer Vision and Pattern Recognition (CVPR)}, June
  2019.

\bibitem{wan2020fine}
J.~Wan, N.~S. Kumar, and A.~B. Chan, ``Fine-grained crowd counting,''
  \emph{arXiv preprint arXiv:2007.06146}, 2020.

\bibitem{Elassal2016Unsupervised}
N.~Elassal and J.~H. Elder, ``Unsupervised crowd counting,'' pp. 329--345,
  2016.

\bibitem{ganin2016domain}
Y.~Ganin, E.~Ustinova, H.~Ajakan, P.~Germain, H.~Larochelle, F.~Laviolette,
  M.~Marchand, and V.~Lempitsky, ``Domain-adversarial training of neural
  networks,'' \emph{Journal of Machine Learning Research}, vol.~17, no.~59, pp.
  1--35, 2016.

\bibitem{tzeng2017adversarial}
E.~Tzeng, J.~Hoffman, K.~Saenko, and T.~Darrell, ``Adversarial discriminative
  domain adaptation,'' in \emph{2017 {IEEE} Conference on Computer Vision and
  Pattern Recognition}, 2017, pp. 2962--2971.

\bibitem{wen2019exploiting}
J.~Wen, R.~Liu, N.~Zheng, Q.~Zheng, Z.~Gong, and J.~Yuan, ``Exploiting local
  feature patterns for unsupervised domain adaptation,'' in \emph{Proceedings
  of the AAAI Conference on Artificial Intelligence}, vol.~33, 2019, pp.
  5401--5408.

\bibitem{wen2019bayesian}
J.~Wen, N.~Zheng, J.~Yuan, Z.~Gong, and C.~Chen, ``Bayesian uncertainty
  matching for unsupervised domain adaptation,'' in \emph{Proceedings of the
  Twenty-Eighth International Joint Conference on Artificial Intelligence,
  {IJCAI-19}}.\hskip 1em plus 0.5em minus 0.4em\relax International Joint
  Conferences on Artificial Intelligence Organization, 2019, pp. 3849--3855.

\bibitem{gao2020nwpu}
Q.~Wang, J.~Gao, W.~Lin, and X.~Li, ``Nwpu-crowd: A large-scale benchmark for
  crowd counting and localization,'' \emph{IEEE Transactions on Pattern
  Analysis and Machine Intelligence}, 2020.

\bibitem{richter2016playing}
S.~R. Richter, V.~Vineet, S.~Roth, and V.~Koltun, ``Playing for data: Ground
  truth from computer games,'' in \emph{European Conference on Computer
  Vision}.\hskip 1em plus 0.5em minus 0.4em\relax Springer, 2016, pp. 102--118.

\bibitem{ros2016synthia}
G.~Ros, L.~Sellart, J.~Materzynska, D.~Vazquez, and A.~M. Lopez, ``The synthia
  dataset: A large collection of synthetic images for semantic segmentation of
  urban scenes,'' in \emph{Proceedings of the IEEE conference on computer
  vision and pattern recognition}, 2016, pp. 3234--3243.

\bibitem{hoffman2016fcns}
J.~Hoffman, D.~Wang, F.~Yu, and T.~Darrell, ``Fcns in the wild: Pixel-level
  adversarial and constraint-based adaptation,'' \emph{arXiv preprint
  arXiv:1612.02649}, 2016.

\bibitem{hoffman2017cycada}
J.~Hoffman, E.~Tzeng, T.~Park, J.-Y. Zhu, P.~Isola, K.~Saenko, A.~A. Efros, and
  T.~Darrell, ``Cycada: Cycle-consistent adversarial domain adaptation,''
  \emph{arXiv preprint arXiv:1711.03213}, 2017.

\bibitem{sankaranarayanan2018learning}
S.~Sankaranarayanan, Y.~Balaji, A.~Jain, S.~Nam~Lim, and R.~Chellappa,
  ``Learning from synthetic data: Addressing domain shift for semantic
  segmentation,'' in \emph{Proceedings of the IEEE Conference on Computer
  Vision and Pattern Recognition}, 2018, pp. 3752--3761.

\bibitem{gao2019weakly}
Q.~Wang, J.~Gao, and X.~Li, ``Weakly supervised adversarial domain adaptation
  for semantic segmentation in urban scenes,'' \emph{IEEE Transactions on Image
  Processing}, vol.~28, no.~9, pp. 4376--4386, 2019.

\bibitem{lee2018diverse}
H.~Lee, H.~Tseng, J.~Huang, M.~Singh, and M.~Yang, ``Diverse image-to-image
  translation via disentangled representations,'' in \emph{Proceedings of the
  IEEE Conference on Computer Vision and Pattern Recognition}, 2018, pp.
  36--52.

\bibitem{chen2019crdoco}
Y.~Chen, Y.~Lin, M.~Yang, and J.~Huang, ``Crdoco: Pixel-level domain transfer
  with cross-domain consistency,'' in \emph{Proceedings of the IEEE Conference
  on Computer Vision and Pattern Recognition}, 2019, pp. 1791--1800.

\bibitem{8784811}
W.~{Li}, L.~{Yongbo}, and X.~{Xiangyang}, ``Coda: Counting objects via
  scale-aware adversarial density adaption,'' in \emph{2019 IEEE International
  Conference on Multimedia and Expo (ICME)}, 2019, pp. 193--198.

\bibitem{zhu2017unpaired}
J.-Y. Zhu, T.~Park, P.~Isola, and A.~A. Efros, ``Unpaired image-to-image
  translation using cycle-consistent adversarial networks,'' \emph{arXiv
  preprint}, 2017.

\bibitem{tsai2018learning}
Y.-H. Tsai, W.-C. Hung, S.~Schulter, K.~Sohn, M.-H. Yang, and M.~Chandraker,
  ``Learning to adapt structured output space for semantic segmentation,'' in
  \emph{Proceedings of the IEEE Conference on Computer Vision and Pattern
  Recognition}, 2018, pp. 7472--7481.

\bibitem{chen2018road}
Y.~Chen, W.~Li, and L.~Van~Gool, ``Road: Reality oriented adaptation for
  semantic segmentation of urban scenes,'' in \emph{Proceedings of the IEEE
  Conference on Computer Vision and Pattern Recognition}, 2018, pp. 7892--7901.

\bibitem{kingma2014adam}
D.~P. Kingma and J.~Ba, ``Adam: A method for stochastic optimization,''
  \emph{arXiv preprint arXiv:1412.6980}, 2014.

\bibitem{gao2019c}
J.~Gao, W.~Lin, B.~Zhao, D.~Wang, C.~Gao, and J.~Wen, ``C$^3$ framework: An
  open-source pytorch code for crowd counting,'' \emph{arXiv preprint
  arXiv:1907.02724}, 2019.

\bibitem{steiner2019pytorch}
B.~Steiner, Z.~Devito, S.~Chintala, S.~Gross, A.~Paszke, F.~Massa, A.~Lerer,
  G.~Chanan, Z.~Lin, E.~Yang \emph{et~al.}, ``Pytorch: An imperative style,
  high-performance deep learning library,'' pp. 8026--8037, 2019.

\bibitem{han2020focus}
T.~Han, J.~Gao, Y.~Yuan, and Q.~Wang, ``Focus on semantic consistency for
  cross-domain crowd understanding,'' in \emph{ICASSP 2020-2020 IEEE
  International Conference on Acoustics, Speech and Signal Processing
  (ICASSP)}.\hskip 1em plus 0.5em minus 0.4em\relax IEEE, 2020, pp. 1848--1852.

\bibitem{wang2004image}
Z.~Wang, A.~C. Bovik, H.~R. Sheikh, and E.~P. Simoncelli, ``Image quality
  assessment: from error visibility to structural similarity,'' \emph{IEEE
  transactions on image processing}, vol.~13, no.~4, pp. 600--612, 2004.

\end{thebibliography}

\end{document}